\documentclass{article}

\usepackage{microtype}
\usepackage{graphicx}
\usepackage{subcaption}
\usepackage{booktabs} 
\usepackage{hyperref}

\usepackage[accepted]{icml2023}

\usepackage{amsmath}
\usepackage{amssymb}
\usepackage{mathtools}
\usepackage{amsthm}

\usepackage[capitalize,noabbrev]{cleveref}

\theoremstyle{plain}
\newtheorem{theorem}{Theorem}[section]

\newtheorem{lemma}[theorem]{Lemma}
\newtheorem{corollary}[theorem]{Corollary}
\theoremstyle{definition}

\theoremstyle{remark}

\usepackage[textsize=tiny]{todonotes}

\icmltitlerunning{Neighborhood Gradient Clustering}

\begin{document}

\twocolumn[
\icmltitle{Neighborhood Gradient Clustering: An Efficient Decentralized Learning Method for Non-IID Data Distributions}



\icmlsetsymbol{equal}{*}

\begin{icmlauthorlist}
\icmlauthor{Sai Aparna Aketi}{equal,yyy}
\icmlauthor{Sangamesh Kodge}{equal,yyy}
\icmlauthor{Kaushik Roy}{yyy}
\end{icmlauthorlist}

\icmlaffiliation{yyy}{Electrical and Computer Engineering, Purdue University, West Lafayette, Indiana, USA}

\icmlcorrespondingauthor{Sai Aparna Aketi}{saketi@purdue.edu}

\icmlkeywords{Machine Learning, ICML}

\vskip 0.3in
]



\printAffiliationsAndNotice{\icmlEqualContribution} 

\begin{abstract}
Decentralized learning algorithms enable the training of deep learning models over large distributed datasets, without the need for a central server. In practical scenarios, the distributed datasets can have significantly different data distributions across the agents. 
This paper focuses on improving decentralized learning over non-IID data with minimal compute and memory overheads. We propose \textit{Neighborhood Gradient Clustering (NGC)}, a novel decentralized learning algorithm that modifies the local gradients of each agent using self- and cross-gradient information. 
In particular, the proposed method replaces the local gradients of the model with the weighted mean of the self-gradients, model-variant cross-gradients 
and data-variant cross-gradients 
The data-variant cross-gradients are aggregated through an additional communication round without breaking the privacy constraints of the decentralized setting. Further, we present \textit{CompNGC}, a compressed version of \textit{NGC} that reduces the communication overhead by $32 \times$ through cross-gradient compression. We theoretically analyze the convergence characteristics of \textit{NGC} and demonstrate its efficiency over non-IID data sampled from {various vision and language} datasets. Our experiments demonstrate that the proposed method either remains competitive or outperforms (by $0-6\%$) the existing state-of-the-art (SoTA) decentralized learning algorithm over non-IID data {with significantly less compute and memory requirements}. Further, we show that the model-variant cross-gradient information available locally at each agent can improve the performance over non-IID data by $1-35\%$ without additional communication cost. 
\end{abstract}

\section{Introduction}

\label{sec:introduction}
The remarkable success of deep learning is mainly attributed to the availability of humongous amounts of data and computing power. 
Large amounts of data are generated on a daily basis at different devices all over the world which could be used to train powerful deep learning models. 
Collecting such data for centralized processing is not practical because of communication and privacy constraints.
To address this concern, a new interest in developing distributed learning algorithms \cite{agarwal2011distributed} has emerged. 
Federated learning (centralized learning) \cite{federated} is a popular setting in the distributed machine learning paradigm, where the training data is kept locally at the edge devices and a global shared model is learned by aggregating the locally computed updates through a coordinating central server. 
Such a setup requires continuous communication with a central server which becomes a potential bottleneck \cite{haghighat2020applications}. 
This has motivated the advancements in decentralized machine learning.

Decentralized machine learning is a branch of distributed learning which focuses on learning from data distributed across multiple agents/devices.
Unlike Federated learning, these algorithms assume that the agents are connected peer to peer without a central server. 
It has been demonstrated that decentralized learning algorithms \cite{d-psgd} can perform comparably to centralized algorithms on benchmark vision datasets.
\cite{d-psgd} present Decentralised Parallel Stochastic Gradient Descent (D-PSGD) by combining SGD with gossip averaging algorithm \cite{gossip}.
Further, the authors analytically show that the convergence rate of D-PSGD is similar to its centralized counterpart \cite{dean2012large}. 
\cite{balu2021decentralized} propose and analyze Decentralized Momentum Stochastic Gradient Descent (DMSGD) which introduces momentum to D-PSGD.
\cite{sgp} introduce Stochastic Gradient Push (SGP) which extends D-PSGD to directed and time-varying graphs.
\cite{deepsqueeze, choco-sgd} explore error-compensated compression techniques (Deep-Squeeze and CHOCO-SGD) to reduce the communication cost of P-DSGD significantly while achieving the same convergence rate as centralized algorithms. 
\cite{sparsepush} combined Deep-Squeeze with SGP to propose communication-efficient decentralized learning over time-varying and directed graphs. 
Recently, \cite{koloskova2020unified} proposed a unified framework for the analysis of gossip-based decentralized SGD methods and provide the best-known convergence guarantees. 

The key assumption to achieve state-of-the-art performance by all the above-mentioned decentralized algorithms is that the data is independent and identically distributed (IID) across the agents. 
In particular, the data is assumed to be distributed in a uniform and random manner across the agents. 
This assumption does not hold in most of the real-world applications as the data distributions across the agent are significantly different (non-IID) based on the user pool \cite{skewscout}. 
The effect of non-IID data in a peer-to-peer decentralized setup is a relatively under-studied problem. 
There are only a few works that try to bridge the performance gap between IID and non-IID data distributions for a decentralized setup. 
Note that, we mainly focus on a common type of non-IID data, widely used in prior works \cite{d2, qgm, cga}: a skewed distribution of data labels across agents.
\cite{d2} proposed $D^2$ algorithm that extends D-PSGD to non-IID data distribution. 
However, the algorithm was demonstrated on only a basic LENET model and is not scalable to deeper models with normalization layers. 
SwarmSGD proposed by \cite{swarmsgd} leverages random interactions between participating agents in a graph to achieve consensus. 
\cite{qgm} replace local momentum with Quasi-Global Momentum (QGM) and improve the test performance by $1-20\%$. 
But the improvement in accuracy is only $1-2\%$ in case of highly skewed data distribution as shown in \cite{lp}. 
Most recently, \cite{cga} proposed Cross-Gradient Aggregation (CGA) and a compressed version of CGA (CompCGA), claiming state-of-the-art performance for decentralized learning algorithms over completely non-IID data. 
CGA aggregates \textit{cross-gradient} information, i.e., derivatives of its model with respect to its neighbors’ datasets through an additional communication round. 
It then updates the model using projected gradients based on quadratic programming. 
CGA and CompCGA require a very slow quadratic programming step \cite{qp} after every iteration for gradient projection which is both compute and memory intensive. 
This work focuses on the following question: \textit{Can we improve the performance of decentralized learning over non-IID data with minimal compute and memory overhead?}

In this paper, we propose \textit{Neighborhood Gradient Clustering} (\textit{NGC}) to handle non-IID data distributions in peer-to-peer decentralized learning setups.
Firstly, we classify the gradients available at each agent into three types, namely self-gradients, model-variant cross-gradients, and data-variant cross-gradients (see Section~\ref{sec:ngc}). 
The self-gradients (or local gradients) are the derivatives computed at each agent on its model parameters with respect to the local dataset.
The model-variant cross-gradients are the derivatives of the received neighbors’ model parameters with respect to the local dataset.
These gradients are computed locally at each agent after receiving the neighbors' model parameters. 
Communicating the neighbors' model parameters is a necessary step in any gossip-based decentralized algorithm \cite{d-psgd}.
The data-variant cross-gradients are the derivatives of the local model with respect to its neighbors’ datasets. 
These gradients are obtained through an additional round of communication.
We then cluster the gradients into a) \textit{model-variant cluster} with self-gradients and model-variant cross-gradients, and b) \textit{data-variant cluster} with self-gradients and data-variant cross-gradients. 
Finally, the local gradients are replaced with the weighted average of the cluster means.
The main motivation behind this modification is to account for the high variation in the computed local gradients (and in turn the model parameters) across the neighbors due to the non-IID nature of the data distribution. 

The proposed technique has two rounds of communication at every iteration to send model parameters and data-variant cross-gradients which incurs $2\times$ communication cost compared to traditional decentralized algorithms (D-PSGD).
To reduce the communication overhead, we propose the compressed version of \textit{NGC} (\textit{CompNGC}) by compressing the additional round of cross-gradient communication. 
Moreover, if the weight associated with the data-variant cluster is set to 0 then \textit{NGC} does not require an additional round of communication.
We provide a detailed convergence analysis of the proposed algorithm and validate the performance of the proposed algorithm on the CIFAR-10 dataset over various model architectures and graph topologies.
We compare the proposed algorithm with D-PSGD, CGA, and CompCGA and show that we can achieve superior performance over non-IID data compared to the current state-of-the-art approach. 
We also report the order of communication, memory, and compute overheads required for \textit{NGC} and CGA as compared to D-PSGD.  

\textbf{Contributions:} In summary, we make the following contributions.
\begin{itemize}
    \item We propose Neighborhood Gradient Clustering (\textit{NGC}) for a decentralized learning setting that utilizes self-gradients, model-variant cross-gradients, and data-variant cross-gradients to improve the learning over non-IID data distribution {(label-wise skew)} among agents.
    \item We theoretically show that the convergence rate of \textit{NGC} is $\mathcal{O}(\frac{1}{\sqrt{NK}}$), which is consistent with the state-of-the-art decentralized learning algorithms.
    \item We present compressed version of Neighborhood Gradient Clustering (\textit{CompNGC}) that reduces the additional round of cross-gradients communication by $32\times$. 
    \item Our experiments show that the proposed method either outperforms by $0-6\%$ or remains competitive with significantly less compute and memory requirements compared to the current state-of-the-art decentralized learning algorithm over non-IID data at iso-communication cost. We also show that when the weight associated with data-variant cross-gradients is set to $0$, \textit{NGC} performs $1-35\%$ better than D-PSGD without any communication overhead.
\end{itemize}

\section{Background}
\label{sec:background}
In this section, we provide the background on decentralized learning algorithms with peer-to-peer connections.

The main goal of decentralized machine learning is to learn a global model using the knowledge extracted from the locally generated and stored data samples across $N$ edge devices/agents while maintaining privacy constraints. In particular, we solve the optimization problem of minimizing global loss function $\mathcal{F}(x)$ distributed across $N$ agents as given in equation.~\ref{eq:1}. 
\begin{equation}
\label{eq:1}
\begin{split}
    \min \limits_{x \in \mathbb{R}^d} \mathcal{F}(x) &= \frac{1}{N}\sum_{i=1}^N f_i(x), \\
    and \hspace{2mm} f_i(x) &= \mathbb{E}_{d^i \in D^i}[F_i(x; d^i)] \hspace{2mm} \forall i
\end{split}
\end{equation}
This is typically achieved by combining stochastic gradient descent \cite{sgd} with global consensus-based gossip averaging \cite{gossip}. 
The communication topology in this setup is modeled as a graph $G = ([N], E)$ with edges $\{i,j\} \in E$ if and only if agents $i$ and $j$ are connected by a communication link exchanging the messages directly. 
We represent $\mathcal{N}(i)$ as the neighbors of $i$ including itself. It is assumed that the graph $G$ is strongly connected with self-loops i.e., there is a path from every agent to every other agent. 
The adjacency matrix of the graph $G$ is referred to as a mixing matrix $W$ where $w_{ij}$ is the weight associated with the edge $\{i,j\}$. Note that, weight $0$ indicates the absence of a direct edge between the agents. 
We assume that the mixing matrix is doubly-stochastic and symmetric, similar to all previous works in decentralized learning. For example, in a undirected ring topology, $w_{ij}=\frac{1}{3}$ if $j \in  \{i-1, i, i+1\}$. 
Further, the initial models and all the hyperparameters are synchronized at the beginning of the training. Algorithm.~\ref{alg:dl} in the appendix describes the flow of D-PSGD with momentum.
The convergence of the Algorithm.~\ref{alg:dl} assumes the data distribution across the agents to be Independent and Identically Distributed (IID). 

\section{Neighborhood Gradient Clustering}
\label{sec:ngc}
We propose the \textit{Neighborhood Gradient Clustering (NGC)} algorithm and a compressed version of \textit{NGC} which improve the performance of decentralized learning over non-IID data distribution. \textit{NGC} utilizes the concepts of self-gradient and cross-gradient \cite{cga}. The following are the definitions of self-gradient and cross-gradient.\\

\textbf{Self-Gradient:} For an agent $i$ with the local dataset $D_i$ and model parameters $x^i$, the self-gradient is the gradient of the loss function $f_i$ with respect to the model parameters $x^i$, evaluated on mini-batch $d^i$ sampled from dataset $D^i$.
\begin{equation}
\label{eq:sg}
\begin{split}
    g^{ii}_{k}=\nabla_x F_i(x^i_{k}; d^i_{k})
\end{split}
\end{equation}
\textbf{Cross-Gradient:} For an agent $i$ with model parameters $x_i$ connected to neighbor $j$ that has local dataset $D^j$, the cross-gradient is the gradient of the loss function $f_j$ with respect to the model parameters $x^i$, evaluated on mini-batch $d^j$ sampled from dataset $D^j$.
\begin{equation}
\label{eq:cg}
\begin{split}
    g^{ij}_{k}=\nabla_x F_j(x^i_{k}; d^j_{k})
\end{split}
\end{equation}

Note that the cross-gradient $g_{ij}$ is computed on agent $j$ using its local data after receiving the model parameters $x_i$ from its neighboring agent $i$ and is then communicated to agent $i$.

\subsection{The \textit{NGC} algorithm}\label{sec:ngc_full}
The flow of the Neighborhood Gradient Clustering (\textit{NGC}) is shown in Algorithm.~\ref{alg:NGC} and the form of the algorithm is similar to D-PSGD \cite{d-psgd} presented in Algorithm.~\ref{alg:dl}.

\begin{algorithm}[ht]
\textbf{Input:} Each agent $i \in [1,N]$ initializes model weights $x^i_{(0)}$, step size $\eta$, momentum coefficient $\beta$, averaging rate $\gamma$, mixing matrix $W=[w_{ij}]_{i,j \in [1,N]}$, \textit{NGC} mixing weight $\alpha$, and $I_{ij}$ are elements of $N\times N$ identity matrix, $\mathcal{N}(i)$ represents neighbors of $i$ including itself.\\

Each agent simultaneously implements the 
T\text{\scriptsize RAIN}( ) procedure\\
1.  \textbf{procedure} T\text{\scriptsize RAIN}( ) \\
2.  \hspace{4mm}\textbf{for} k=$0,1,\hdots,K-1$ \textbf{do}\\
3.  \hspace*{8mm}$d^i_{k} \sim D^i$\\
4.  \hspace*{8mm}$g^{ii}_{k}=\nabla_x f_i(d^i_{k}; x^i_{k}) $ \\
5.  \hspace*{8mm}S\text{\scriptsize END}R\text{\scriptsize ECEIVE}($x^i_{t}$)\\
6.  \hspace*{8mm}\textbf{for} each neighbor $j \in \{\mathcal{N}(i)-i\}$ \textbf{do}\\
7.  \hspace*{12mm}$g^{ji}_{k}=\nabla_x f_i(d^i_{k}; x^j_{k})$\\
8.  \hspace*{12mm}\textbf{if} $\alpha \neq 0$ \textbf{do}\\
9.  \hspace*{16mm}S\text{\scriptsize END}R\text{\scriptsize ECEIVE}($g^{ji}_{k}$)\\
10.\hspace*{12mm}\textbf{end}\\
11. \hspace*{8mm}\textbf{end}\\
12. \hspace*{8mm}$\widetilde{g}^i_{k}=\sum_{j\in \mathcal{N}(i)}[(1-\alpha)w_{ji}g^{ji}_{t}+ \alpha w_{ij}g^{ij}_{t}]$\\
13. \hspace*{8mm}$v^i_{k}= \beta v^i_{(k-1)} - \eta \widetilde{g}^i_{k}$\\
14. \hspace*{8mm}$\widetilde{x}^i_{k}=x^i_{k}+v^i_{k}$\\
15. \hspace*{8mm}$x^i_{(k+1)}=\widetilde{x}^i_{t}+\gamma\sum_{j\in \mathcal{N}(i)}(w_{ij}-I_{ij})x^j_{k}$\\
16. \hspace{4mm}\textbf{end}\\
17. \textbf{return}
\caption{Neighborhood Gradient Clustering (\textit{NGC})}
\label{alg:NGC}
\end{algorithm}

The main contribution of the proposed \textit{NGC} algorithm is the local gradient manipulation step (line 12 in Algorithm.~\ref{alg:NGC}). 
In the $k^{th}$ iteration of \textit{NGC}, each agent $i$ calculates its self-gradient $g^{ii}$. 
Then, agent $i$’s model parameters are transmitted to all other agents ($j$) in its neighborhood, and the respective cross-gradients are calculated by the neighbors and transmitted back to agent $i$.
At every iteration after the communication rounds, each agent $i$ has access to self-gradients ($g^{ii}$) and two sets of cross-gradients: 
1) \textit{Model-variant cross-gradients}: The derivatives that are computed locally using its local data on the neighbors' model parameters ($g^{ji}$). 
2) \textit{Data-variant cross-gradients}: The derivatives (received through communication) of its model parameters on the neighbors' dataset ($g^{ij}$). 
Note that each agent $i$ computes and transmits cross-gradients ($g^{ji}$) that act as model-variant cross-gradients for $i$ and as data-variant cross-gradients for $j$.
We then cluster the gradients into two groups namely: a) \textit{Model-variant cluster} $\{g^{ji} \forall j \in \mathcal{N}(i)\}$ that includes self-gradients and model-variant cross-gradients, and b) \textit{Data-variant cluster} $\{g^{ij} \forall j \in \mathcal{N}(i)\}$ that includes self-gradients and data-variant cross-gradients. 
The local gradients at each agent are replaced with the weighted average of the above-defined cluster means as shown in Equation.~\ref{eq:ngc}, which assumes uniform mixing matrix ($w_{ij}=1/m; m = |\mathcal{N}(i)|$).
The mean of the model-variant cluster is weighted by $(1-\alpha)$ and the mean of the data-variant cluster is weighed by $\alpha$ where $\alpha \in [0,1]$ is a hyper-parameter referred to as \textit{NGC} mixing weight.
\begin{equation}
\label{eq:ngc}
\begin{split}
    \widetilde{g}^i_{k}&= (1-\alpha)*\underbrace{\Bigr[\frac{1}{m}\sum_{j\in \mathcal{N}(i)}g^{ji}_{k}\Bigr]}_\text{\hspace{-10mm}\clap{\textbf{(a) Model-variant cluster mean}~}} + \alpha*\underbrace{\Bigr[\frac{1}{m}\sum_{j\in \mathcal{N}(i)}g^{ij}_{k}\Bigr]}_\text{\clap{\textbf{(b) Data-variant cluster mean}~}}\\ 
\end{split}
\end{equation}

The motivation for this modification is to reduce the variation of the computed local gradients across the agents. In IID settings, the local gradients should statistically resemble the cross-gradients and hence simple gossip averaging is sufficient to reach convergence. However, in the non-IID case, the local gradients across the agents are significantly different due to the variation in datasets and hence the model parameters on which the gradients are computed. 
The proposed algorithm reduces this variation in the local-gradients as it is equivalent to adding two bias terms $\epsilon$ and $\omega$ with weights ($1-\alpha$) and $\alpha$ respectively as shown in Equation.~\ref{eq:ngc_dual}.
\begin{equation}
\label{eq:ngc_dual}
\begin{split}
    \widetilde{g}^i_{k}
    &= g^{ii}_{k} + (1-\alpha)*\underbrace{\Bigr[\frac{1}{m}\sum_{j\in \mathcal{N}(i)}(g^{ji}_{k}-g^{ii}_{k})\Bigr]}_{\text{model variance bias}\hspace{1mm}\epsilon^{i}_{k}}\\
    &\hspace{10mm}+  \alpha*\underbrace{\Bigr[\frac{1}{m}\sum_{j\in \mathcal{N}(i)} \frac{1}{m}(g^{ij}_{k}-g^{ii}_{k})\Bigr]}_{\text{data variance bias}\hspace{1mm}\omega^{i}_{k}}\\
     \epsilon^i_k&=  \frac{1}{m}*\sum_{j\in \mathcal{N}(i)}\big(\nabla_x F(d^i_{k}; x^j_{k})-\nabla_x F(d^i_{k}; x^i_{k})\big)\\
    \omega^i_k&=\frac{1}{m}*\sum_{j\in \mathcal{N}(i)}\big(\nabla_x F(d^j_{k}; x^i_{k})-\nabla_x F(d^i_{k}; x^i_{k})\big)\\
\end{split}
\end{equation}
The bias term $\epsilon$ compensates for the difference in a neighborhood's self-gradients caused due to variation in the model parameters across the neighbors. Whereas, the bias term $\omega$ compensates for the difference in a neighborhood's self-gradients caused due to variation in the data distribution across the neighbors. We hypothesize and show through our experiments that the addition of these bias terms to the local gradients improves the performance of decentralized learning over non-IID data by accelerating global convergence. Note that if we set $\alpha = 0$ in the \textit{NGC} algorithm then it does not require an additional communication round (no communication overhead compared to D-PSGD). 

\subsection{The Compressed \textit{NGC} Algorithm} \label{sec:compngc}
The \textit{NGC} algorithm at every iteration involves two rounds of communication with the neighbors: 1) communicate the model parameters, and 2) communicate the cross-gradients. This communication overhead can be a bottleneck in a resource-constrained environment. Hence we propose a compressed version of \textit{NGC} using Error Feedback SGD (EF-SGD) \cite{karimireddy2019error} to compress gradients. We compress the error-compensated self-gradients and cross-gradients from 32 bits (floating point precision of arithmetic operations) to 1 bit by using scaled signed gradients. The error between the compressed and non-compressed gradient of the current iteration is added as feedback to the gradients in the next iteration before compression. The pseudo-code for \textit{CompNGC} is shown in Algorithm.~\ref{apx_alg:compNGC} in the Appendix.

\section{Convergence Analysis of \textit{NGC}} 
\label{sec:convergence}

In this section, we provide a convergence analysis for \textit{NGC} Algorithm. We assume that the following statements hold:

\textbf{Assumption 1 - Lipschitz Gradients:} Each function $f_i(x)$ is L-smooth.

\textbf{Assumption 2 - Bounded Variance:} The variance of the stochastic gradients is assumed to be bounded.
\begin{equation}
\label{eq:inner_variance}
\mathbb{E}_{ d \sim D_i} || \nabla F_i(x; d) - \nabla f_i(x)||^2 \leq \sigma^2 \hspace{2mm} \forall i \in [1,N]
\end{equation}
\begin{equation}
\label{eq:outer_variance}
  || \nabla f_i(x) - \nabla \mathcal{F}(x)||^2 \leq \zeta^2 \hspace{2mm} \forall i \in [1,N]
\end{equation}
\textbf{Assumption 3 - Doubly Stochastic Mixing Matrix:} The mixing matrix $W$ is a real doubly stochastic matrix with $\lambda_1(W)=1$ and
\begin{equation}
\label{eq:eigen}
max{\{|\lambda_2(W)|, |\lambda_N(W)|\}} \leq \sqrt{\rho}<1
\end{equation}
where $\lambda_i(W)$ is the $i^{th}$ largest eigenvalue of W and $\rho$ is a constant.\\
The above assumptions are commonly used in most decentralized learning setups. 
\begin{lemma} \label{lemma1}
Given assumptions 1-3, we define  $g_i = \nabla F_i(x; d)$ and the \textit{NGC} gradient update $\widetilde{g}_i$. For all $i$, we have:
\begin{equation}
\label{eq:lemma1}
    \mathbb{E}\bigg[\bigg\|\frac{1}{N}\sum_{i=1}^N(\tilde{\mathbf{g}}^i -\mathbf{g}^i)\bigg\|^2\bigg] \leq 4 (\frac{\sigma^2}{N}+\zeta^2)
\end{equation}
\end{lemma}
A complete proof for Lemma~\ref{lemma1} can be found in Appendix.~\ref{apx:lemma1}.  The gradient variation bounded by $\zeta^2$ determines the heterogeneity in the data distribution i.e., the non-IIDness and Lemma~\ref{lemma1} shows that the \textit{NGC} gradient update is bounded by $\zeta^2$. Intuitively, the distance between the self-gradients $g^i$ and the \textit{NGC} gradient update $\widetilde{g}^i$ increases with an increase in the degree of heterogeneity in data distribution which is expected. 

Theorem~\ref{theorem_1} presents the convergence of the proposed algorithm and the proof is detailed in Appendix~\ref{apx:theorem_1}

\begin{theorem}\label{theorem_1}
(Convergence of \textit{NGC} algorithm) Given assumptions 1-3  and lemma~\ref{lemma1}, let step size $\eta$ satisfies the following conditions:
\begin{equation}
\label{eq:eta}
\begin{split}
     1-\frac{6\eta^2 L^2}{(1-\beta)^2(1-\sqrt{\rho})}-\frac{4L\eta}{(1-\beta)^2} \geq 0
\end{split}
\end{equation}
For all $K \geq 1$, we have
\begin{equation}
    \begin{split}
        \frac{1}{K}&\sum_{k=0}^{K-1}  \mathbb{E}[||\nabla  \mathcal{F}(\Bar{x}_k)||^2] \leq \frac{1}{C_1K}(\mathcal{F}(\Bar{x}^0)-\mathcal{F}^*) +\\
        &\hspace{-2mm}\bigg(10C_{2}+10C_3\frac{\eta^2\beta^2}{(1-\beta)^4} + 4C_4\bigg)(\frac{\sigma^2}{N}+\zeta^2)+ \\
        &\hspace{-2mm} C_5 \frac{14\eta^2(\sigma^2+\zeta^2)}{(1-\beta)^2(1-\sqrt{\rho})^2}  
    \end{split}
\end{equation}
where $\beta$ = momentum coefficient, $\Bar{x}_k = \frac{1}{N} \sum_{i=1}^{N}x^i_k$\\
$C_1 = \frac{1}{2} (\frac{\eta}{1-\beta}-\frac{(1-\beta)}{\beta L})$,\\ 
$C_{2}=\left(\frac{\beta L \eta^{2}}{2(1-\beta)^{3}}+\frac{\eta^{2} L}{(1-\beta)^{2}}\right) / C_{1}$, 
$C_{3}=\frac{(1-\beta) L}{2 \beta} / C_{1}$,\\
$C_{4}=\frac{\eta^2 \beta L}{2(1-\beta)^{3}} / C_{1}$, and 
$C_{5}=\frac{\eta L^{2}}{2(1-\beta)} / C_{1}$.

\end{theorem}

The result of the theorem.~\ref{theorem_1} shows that the magnitude of the average gradient achieved by the consensus model is upper-bounded by the difference between the initial objective function value and the optimal value, the sampling variance, and gradient variations across the agents representing data heterogeneity. A detailed explanation of the constraints on step size is presented in the Appendix.~\ref{step-size}. Further, we present a corollary to show the convergence rate of \textit{NGC} in terms of the number of iterations.

\begin{corollary}
\label{corol}
Suppose that the step size satisfies $\eta=\mathcal{O}\Big(\sqrt{\frac{N}{K}}\Big)$ and that $\zeta^2=\mathcal{O}\Big(\frac{1}{\sqrt{K}}\Big)$.  
For a sufficiently large $K\geq \frac{36NL^2}{r^2}, and \\
r=\sqrt{4(1-\sqrt{\rho})^2+6(1-\sqrt{\rho})(1-\beta)^2}-2(1-\sqrt{\rho})$, we have, for some constant $C > 0$, 
\begin{align}
        \frac{1}{K}\sum_{k=0}^{K-1} \mathbb{E}&\left[\left\|\nabla \mathcal{F}\left(\bar{\mathbf{x}}_k\right)\right\|^{2}\right] \nonumber 
        \leq C\Bigg(\frac{1}{\sqrt{NK}}+\frac{1}{K} +\frac{1}{K^{1.5}}
        \Bigg).
\end{align}
\end{corollary}

The proof for Corollary.~\ref{corol} can be found in Appendix.~\ref{coro_1_proof}. It indicates that the proposed algorithm achieves linear speedup (with a convergence rate of $\mathcal{O}(\frac{1}{\sqrt{NK}}))$ when $K$ is sufficiently large. This convergence rate is similar to the well-known best result for decentralized SGD algorithms in the literature as shown in Table.~\ref{tab:rates}. 

\begin{table}[ht]
\caption{Convergence rate comparison between decentralized learning algorithms.}
\label{tab:rates}
\vspace{-1mm}
\begin{center}
\begin{tabular}{lc}
\hline
Method & Rate\\
\hline
D-PSGD & $\mathcal{O}(\frac{1}{\sqrt{NK}} + \frac{1}{K} )$ \\
SwarmSGD  & $\mathcal{O}(\frac{1}{\sqrt{K}})$ \\
CGA & $\mathcal{O}(\frac{1}{\sqrt{NK}} + \frac{1}{K} + \frac{1}{K^{1.5}} + \frac{1}{K^{2}})$ \\
\textit{NGC} (ours) & $\mathcal{O}(\frac{1}{\sqrt{NK}} + \frac{1}{K} + \frac{1}{K^{1.5}})$ \\
\hline
\end{tabular}
\end{center}
\end{table}

\begin{table*}[h]
\caption{Average test accuracy comparisons for CIFAR-10 with non-IID data using various architectures and graph topologies. The results are averaged over three seeds where std is indicated.}
\label{tab:cf10}
\begin{center}
\begin{tabular}{l c c c c c}
\hline
Method& Agents& 5layer CNN& 5layer CNN&VGG-11 & ResNet-20\\
& & Ring & Torus & Ring & Ring \\
 \hline
 \hline
     &  5 &  76.00 $\pm$ 1.44&-& 67.04 $\pm$ 5.36&82.13 $\pm$ 0.84\\
     D-PSGD &  10 & 47.68 $\pm$ 3.20& 55.34 $\pm$ 6.32&44.14 $\pm$ 3.30 &31.66 $\pm$ 6.01\\
     &  20 & 44.85 $\pm$ 1.94& 50.12 $\pm$ 1.91 &38.92 $\pm$ 2.99 & 31.94 $\pm$ 2.91\\
     \hline
      &  5 & \textbf{82.20 $\pm$ 0.34}& -& \textbf{ 79.27 $\pm$ 0.39}&\textbf{85.88 $\pm$ 0.58}\\
    \textit{NGC} (ours) &  10 &\textbf{67.43 $\pm$ 1.15} &\textbf{73.84 $\pm$ 0.33} &\textbf{ 59.92 $\pm$  2.12} &\textbf{66.02 $\pm$ 2.86}\\
    ($\alpha=0$) &  20 & \textbf{58.80 $\pm$ 1.30} &\textbf{64.55 $\pm$ 1.16} & \textbf{ 52.70$\pm$ 1.63}&\textbf{50.74 $\pm$ 2.36}\\
         \hline
  \hline
     &  5 &82.20 $\pm$ 0.43 & -& 84.41 $\pm$ 0.22&87.52 $\pm$ 0.50\\
     CGA &  10 & 72.96 $\pm$ 0.40 &76.04 $\pm$ 0.62 & \textbf{ 79.66 $\pm$ 0.46 }& 79.98 $\pm$ 1.23\\
     &   20 &69.88 $\pm$ 0.84& 73.21 $\pm$ 0.27 &79.30 $\pm$ 0.12 &75.13 $\pm$ 1.56 \\
     \hline
 &  5 &\textbf{83.36 $\pm$ 0.65} &- &\textbf{85.15 $\pm$ 0.58} &\textbf{88.52 $\pm$ 0.19}\\
    \textit{NGC} (ours) &  10 &\textbf{75.34 $\pm$ 0.30} &\textbf{78.53 $\pm$ 0.56} &79.55 $\pm$ 0.30 & \textbf{84.02 $\pm$ 0.44}\\
   &  20 & \textbf{73.36 $\pm$ 0.88} &\textbf{75.11 $\pm$ 0.07} & \textbf{79.43 $\pm$ 0.62}&\textbf{81.26 $\pm$ 0.69}\\
   \hline
\hline
   &   5 &82.00 $\pm$ 0.25 & -& 83.65 $\pm$ 0.41&86.73 $\pm$ 0.34\\
    CompCGA &  10 & 71.41 $\pm$ 0.94 &75.95 $\pm$ 0.41 & 73.96 $\pm$ 0.31 & 73.63 $\pm$ 0.55\\
     &   20 & 68.15 $\pm$ 0.79& 71.71 $\pm$ 0.54& 73.72 $\pm$ 2.74&66.34 $\pm$ 0.98\\
 \hline
  &  5 & \textbf{82.91 $\pm$ 0.21} & -&\textbf{ 84.03 $\pm$ 0.32} &\textbf{87.56 $\pm$ 0.34}\\
   \textit{CompNGC} (ours) & 10 & \textbf{74.36 $\pm$ 0.42}& \textbf{77.82 $\pm$ 0.20}&\textbf{ 77.02 $\pm$ 0.14 } & \textbf{78.50 $\pm$ 0.98}\\
     &  20 &\textbf{71.46 $\pm$ 0.85}&\textbf{73.62 $\pm$ 0.74} &\textbf{ 73.76 $\pm$ 0.20 } &  \textbf{72.62 $\pm$ 0.71}\\
 \hline
\end{tabular}
\end{center}
\end{table*}

\begin{table*}[h]
\caption{{Average test accuracy comparisons for various datasets with non-IID sampling trained over undirected ring topology. The results are averaged over three seeds where std is indicated.}}
\label{tab:datasets}
\begin{center}
\begin{tabular}{l c c c c }
\hline
Method& Agents& Fashion MNIST & CIFAR-100 & Imagenette \\
& & (LeNet-5) & (ResNet-20) & (MobileNet-V2)  \\
 \hline
 \hline
       &  5 & 86.43 $\pm$ 0.14 & 44.66 $\pm$ 5.23 & 47.09 $\pm$ 9.20 \\
   D-PSGD  &  10 & 75.49 $\pm$ 0.32 & 19.03 $\pm$ 13.27 & 32.81 $\pm$ 2.18\\
     \hline
    \textit{NGC} (ours) &  5 &  \textbf{88.49 $\pm$ 0.18} & \textbf{55.96 $\pm$ 0.95} & \textbf{60.15 $\pm$ 2.17}\\
    ($\alpha=0$)   &  10 & \textbf{82.85 $\pm$ 0.24}  & \textbf{35.34 $\pm$ 0.32} & \textbf{36.13 $\pm$ 1.97}\\
  \hline
  \hline
     &  5 & 90.03 $\pm$ 0.39  &56.43  $\pm$ 2.39 & 72.82 $\pm$ 1.25 \\
    CGA  &  10 & \textbf{87.61 $\pm$ 0.30}  & 53.61 $\pm$ 1.07 & 61.97 $\pm$ 0.58\\
     \hline
  &  5 & \textbf{90.61 $\pm$ 0.18}  & \textbf{56.50 $\pm$ 3.23} & \textbf{74.49 $\pm$ 0.93}\\
    \textit{NGC} (ours) &  10 & 87.24 $\pm$ 0.23 & \textbf{53.77 $\pm$ 0.15}   & \textbf{64.06 $\pm$ 1.11} \\
  \hline
  \hline
     &   5 & 90.45 $\pm$ 0.34&55.74 $\pm$ 0.33  & 72.76 $\pm$ 0.44\\
   CompCGA &  10 & 81.62 $\pm$ 0.37 & 38.84 $\pm$ 0.54 & 59.92 $\pm$ 0.72\\
 \hline
   &  5 & \textbf{90.48 $\pm$ 0.19} & \textbf{57.51 $\pm$ 0.48} & \textbf{72.91 $\pm$ 1.06}\\
  \textit{CompNGC} (ours)  & 10 & \textbf{83.38 $\pm$ 0.39} & \textbf{43.07 $\pm$ 0.32}& \textbf{61.91 $\pm$ 2.10}\\
 \hline
\end{tabular}
\end{center}
\end{table*}

\begin{table*}[h]
\caption{{Average test accuracy comparisons for AGNews dataset (left side of the table) and full resolution ($224 \times 224$) Imagenette dataset (right side of the table). The results are averaged over three seeds where std is indicated.}}
\label{tab:nlp-fr_imagenette}
\begin{center}
\begin{tabular}{l  c  c  c || c c }
\hline
Method& \multicolumn{2}{c}{AGNews-BERT\textsubscript{mini} } & AGNews-DistilBERT\textsubscript{base}  & \multicolumn{2}{c}{Imagenette-ResNet-18 }\\
& Agents = 4 &Agents = 8  & Agents = 4&\multicolumn{2}{c}{Agents = 5}\\
& \multicolumn{2}{c}{Ring Topology} & Ring Topology& Ring Topology&Chain Topology\\
 \hline
 \hline
    D-PSGD & 89.21 $\pm$ 0.41   &85.48 $\pm$ 0.71& 91.54 $\pm$ 0.07 & 65.43 $\pm$ 4.60 & 42.02 $\pm$ 1.25\\
    \textit{NGC} $\alpha=0$ (ours)&  \textbf{89.40 $\pm$ 0.13} &\textbf{87.58 $\pm$ 0.07} & \textbf{91.70 $\pm$ 0.11}&  \textbf{73.15 $\pm$ 0.38}  & \textbf{47.87 $\pm$ 0.99} \\
    \hline
    CGA &  91.43 $\pm$ 0.11&\textbf{89.15 $\pm$ 0.45} & 93.42 $\pm$ 0.04 &  85.00 $\pm$ 0.67 & 65.96 $\pm$ 1.84 \\
    \textit{NGC} (ours) &  \textbf{92.24 $\pm$ 0.29}  &89.02 $\pm$ 0.39 & \textbf{94.11 $\pm$ 0.01}&  \textbf{85.85 $\pm$ 0.60}  & \textbf{67.77 $\pm$ 1.76} \\
      \hline
    CompCGA &   91.05 $\pm$ 0.29   & 88.91 $\pm$ 0.25 & \textbf{93.54 $\pm$ 0.03}&   84.65 $\pm$ 0.57 & \textbf{62.93 $\pm$ 1.33} \\
    \textit{CompNGC} (ours)&  \textbf{91.24 $\pm$ 0.43}  &\textbf{ 89.01 $\pm$ 0.13}& 93.50 $\pm$ 0.16&  \textbf{85.44 $\pm$ 0.10} & 62.64 $\pm$ 0.85\\
     \hline
\end{tabular}
\end{center}
\end{table*}
\section{Experiments}\label{sec:results}
In this section, we analyze the performance of the proposed \textit{NGC} and \textit{CompNGC} techniques and compare them with the baseline D-PSGD algorithm \cite{d-psgd} and state-of-the-art CGA and CompCGA methods \cite{cga}. \footnote{Our PyTorch code is available at \url{https://github.com/aparna-aketi/neighborhood_gradient_clustering}}

\emph{\textbf{Experimental Setup:}} {The efficiency of the proposed method is demonstrated through our experiments on a diverse set of datasets, model architectures, tasks, topologies, and numbers of agents}. {We present the analysis on -- 
(a) Datasets (Appendix~\ref{apx:datasets}): vision datasets (CIFAR-10, CIFAR-100, Fashion MNIST and Imagenette \cite{imagenette}) and language datasets (AGNews \cite{agnews}}).
{(b) Model architectures (Appendix~\ref{apx:arch}):  5-layer CNN, VGG-11, ResNet-20, LeNet-5, MobileNet-V2, ResNet-18, BERT\textsubscript{mini} and DistilBERT\textsubscript{base} }
{(c) Tasks: Image and Text Classification.}
{(d) Topologies: Ring, Chain, and Torus.}
{(e) Number of agents: varying from 4 to 20.}
{Note that we use low resolution ($32 \times 32$) images of Imagenette dataset for the experiments in Table.~\ref{tab:datasets}. The results for high resolution ($224 \times 224$) Imagenette are presented in Table.~\ref{tab:nlp-fr_imagenette}.}
We consider an extreme case of non-IID distribution where no two neighboring agents have the same class. This is referred to as complete label-wise skew or 100\% label-wise non-IIDness \cite{skewscout}. 
In particular, for a 10-class dataset such as CIFAR-10 - each agent in a 5-agent system has data from 2 distinct classes, and each agent in a 10 agents system has data from a unique class.
For a 20 agent system two agents that are maximally apart share the samples belonging to a class.
{We report the test accuracy of the consensus model averaged over three randomly chosen seeds.
The details of the hyperparameters for all the experiments are present in Appendix.~\ref{apx:hyperparameters}.  We compare the proposed method with iso-communication baselines. The experiments on \textit{NGC} ($\alpha=0$) are compared with D-PSGD, \textit{NGC} with CGA, and \textit{CompNGC} with CompCGA. The communication cost for each experiment in this section is presented in Appendix~\ref{apx:comm}.}
\begin{figure*}[h]
\centering
\begin{subfigure}{0.30\textwidth}
    \vspace{-4mm}
    \includegraphics[width=\textwidth]{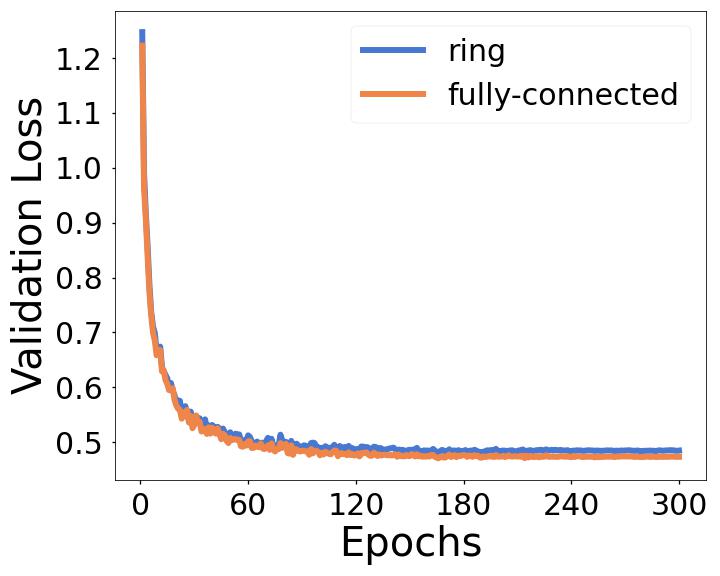}
    \caption{\textit{NGC} method on IID data.}
    \label{fig:iid_curve}
\end{subfigure}
\hfill
\begin{subfigure}{0.30\textwidth}
 \vspace{-4mm}
    \includegraphics[width=\textwidth]{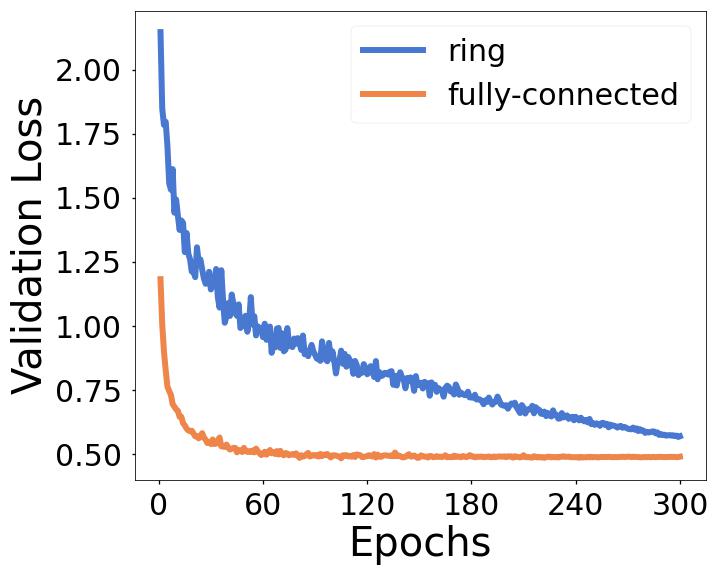}
    \caption{\textit{NGC} method on Non-IID data.}
    \label{fig:noniid_curve}
\end{subfigure}
\hfill
\begin{subfigure}{0.30\textwidth}
    \includegraphics[width=\textwidth]{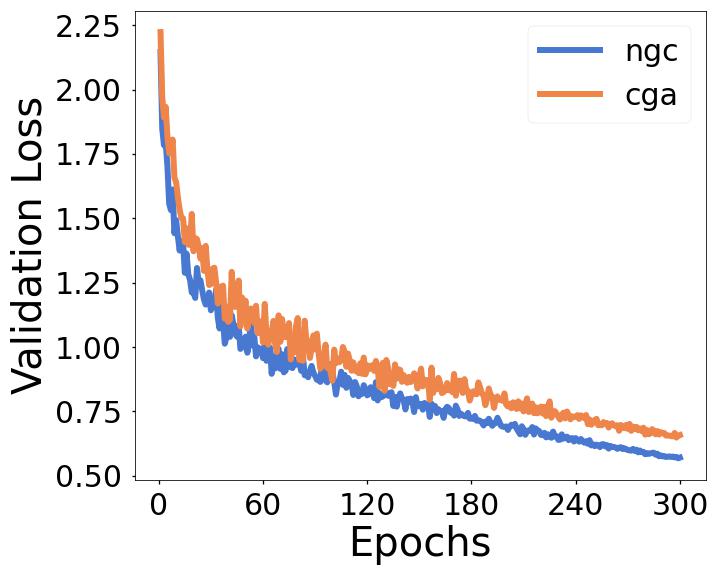}
    \caption{Different methods on Non-IID for ring topology.}
    \label{fig:algo_curve}
\end{subfigure}        
\caption{Average Validation loss during training of 5 agents on CIFAR-10 with a 5 layer CNN.}
\label{fig:figures}
\end{figure*}

\emph{\textbf{Results:}}
{We evaluate variants of \textit{NGC} and \textit{CompNGC} and compare them with respective baselines in Table.~\ref{tab:cf10}, for training different models trained on CIFAR-10 over various graph sizes and topologies.
We observe that \textit{NGC} with $\alpha=0$ consistently outperforms D-PSGD for all models, graph sizes, and topologies with a significant performance gain varying from $3-35\%$. Our experiments show the superiority of \textit{NGC} and \textit{CompNGC} over CGA and CompCGA respectively. The performance gains are more pronounced when considering larger graphs (with 20 agents) and compact models such as ResNet-20. We further demonstrate the generalizability of the proposed method by evaluating it on various image datasets such as Fashion MNIST, Imagenette and on challenging dataset such as CIFAR-100. Table.~\ref{tab:datasets}, \ref{tab:nlp-fr_imagenette} show that \textit{NGC} with $\alpha=0$ outperforms D-PSGD by $2-13\%$ across various datasets where as \textit{NGC} and \textit{CompNGC} remain competitive with an average improvement of $\sim 1\%$. 
To show the effectiveness of the proposed method across different modalities, we present results on the text classification task in Table~\ref{tab:nlp-fr_imagenette}. 
We train on BERT\textsubscript{mini} model with AGNews dataset distributed over 4 and 8 agents and a larger transformer model (DistilBert\textsubscript{base}) distributed over 4 agents. 
For NGC $\alpha=0$ we see a maximum improvement of $2.1$\% over the baseline D-PSGD algorithm. 
Even for the text classification task, we observe \textit{NGC} and \textit{CompNGC} to be competitive with CGA and CompCGA methods.
These observations are consistent with the results on the image classification tasks.
Finally, through this exhaustive set of experiments, we demonstrate that the weighted averaging of data-variant and model-variant cross-gradients can be served as a simple plugin to boost the performance of decentralized learning over label-wise non-IID data. Further, locally available model-variant cross-gradients information at each agent can be efficiently utilized to improve decentralized learning with no communication overhead.} 

\begin{figure*}[h]
\minipage{0.33\textwidth}
  \includegraphics[width=\linewidth]{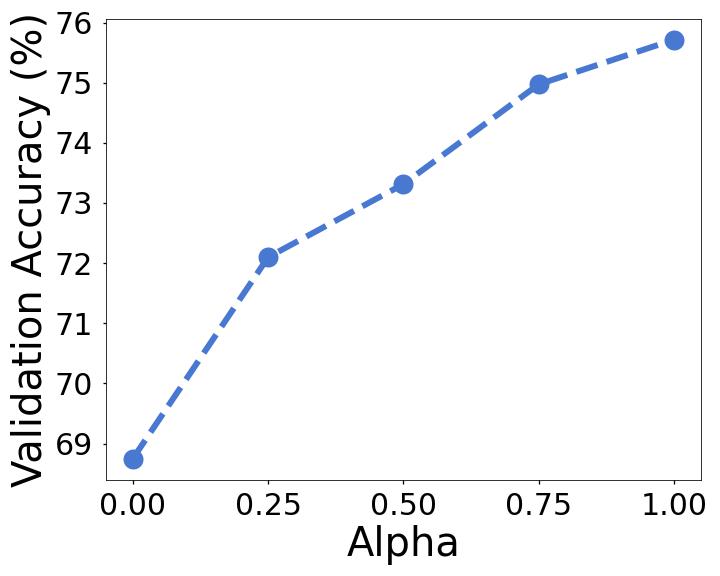}
  \caption{\textit{NGC} mixing weight $\alpha$ variation for 10 agents trained on ring graph with 5 layer CNN.}\label{fig:alpha}
\endminipage\hfill
\minipage{0.63\textwidth}
     \centering
     \begin{subfigure}[htbp]{0.49\linewidth}
         \centering
         \includegraphics[width=\linewidth]{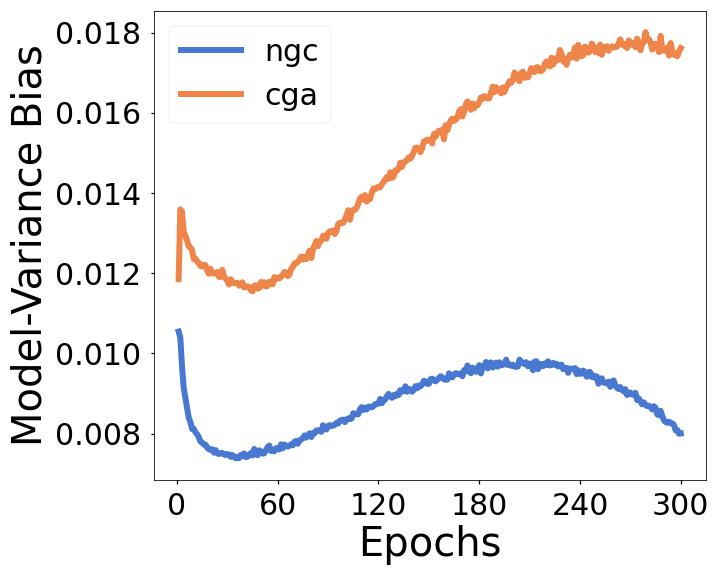}
         \caption{Average L1 norm of $\epsilon$.}
         \label{fig:epsilon}
     \end{subfigure}
     \begin{subfigure}[htbp]{0.49\linewidth}
         \centering
         \includegraphics[width=\linewidth]{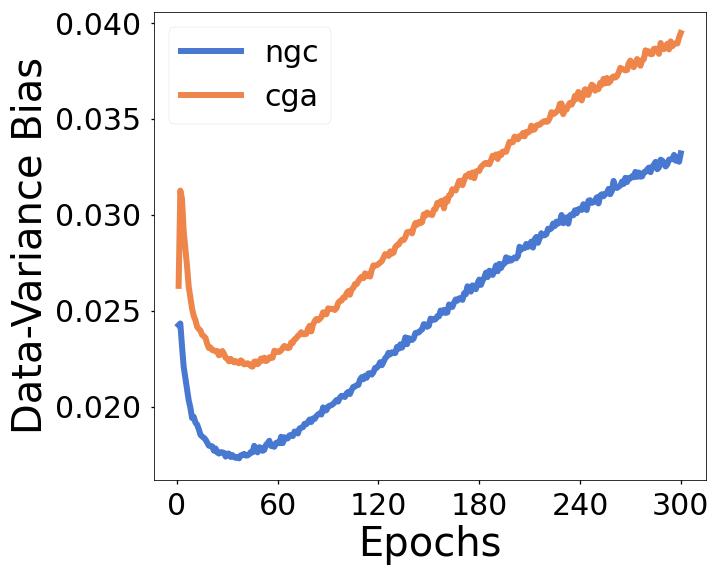}
         \caption{Average L1 norm of $\omega$.}
         \label{fig:omega}
     \end{subfigure}
    \caption{ Average L1 norm of model variance bias and data variance bias for 5 agents trained on ring graph with 5 layer CNN.}
        \label{fig:norms}
\endminipage
\end{figure*}

{\textbf{\emph{Analysis:}}} We empirically show the convergence characteristics of the proposed algorithm over IID and Non-IID data distributions in Figure.~\ref{fig:iid_curve}, and \ref{fig:noniid_curve} respectively. For Non-IID distribution, we observe that there is a slight difference in convergence (as expected) with a slower rate for sparser topology (ring graph) compared to denser counterpart (fully connected graph). Figure.~\ref{fig:algo_curve} shows the comparison of the convergence characteristics of the \textit{NGC} algorithm with the current state-of-the-art CGA algorithm. We observe that \textit{NGC} has lower validation loss than CGA for the same decentralized setup. Analysis for 10 agents is presented in Appendix~\ref{apx:10agent}.
The change in average validation accuracy with \textit{NGC} mixing weight $\alpha$ is shown in Figure.~\ref{fig:alpha}.
We observe that $\alpha$ close to 1 has the best performance as the model variance bias is taken care in the gossip averaging step.
We also plot the model variance and data variance bias terms for both \textit{NGC} and CGA techniques as shown in Figure.~\ref{fig:epsilon}, and \ref{fig:omega} respectively. 
We observe that both the model variance and the data variance bias for \textit{NGC} are significantly lower than CGA. 
This is because CGA gives more importance to self-gradients as it updates in the descent direction that is close to self-gradients and is positively correlated to data-variant cross-gradients. 
In contrast, \textit{NGC} accounts for the biases directly and gives equal importance to self-gradients and data-variant cross-gradients, thereby achieving superior performance.

{\textbf{\emph{Hardware Benefits:}}} The proposed \textit{NGC} algorithm is superior in terms of memory and compute efficiency (see Table.~\ref{apx_tab:hardware} in Appendix.~\ref{apx:hardware}) while having equal communication cost as compared to \textit{CGA}. 
Since \textit{NGC} involves weighted averaging, an additional buffer to store the cross-gradients is not required. Weighted cross-gradients can be added to the self-gradient buffer. 
CGA stores all the cross-gradients in a matrix form for quadratic programming projection of the local gradient. 
Therefore, \textit{NGC} has no memory overhead compared to the baseline D-PSGD algorithm, while CGA requires additional memory equivalent to the number of neighbors times model size.
Moreover, the quadratic programming projection step \cite{qp} in CGA is much more expensive in terms of compute and latency as compared to the weighted averaging step of cross-gradients in \textit{NGC}. 
Our experiments clearly show that \textit{NGC} is superior to CGA in terms of test accuracy, memory efficiency, compute efficiency, and latency.

\begin{table}[h]
\caption{Comparison of communication, memory, and compute overheads per mini-batch compared to D-PSGD. $m_s$: model size, $N_i$: number of neighbors, $b$: floating point precision, Q: compute for Quadratic Programming, B: compute for Backward Pass.}
\label{apx_tab:hardware}
\begin{center}
\resizebox{1.0\columnwidth}{!}{
\begin{tabular}{lccc}
\hline
Method & Comm. & Memory & Compute\\
\hline
CGA &$\mathcal{O}(m_s N_i)$ &$\mathcal{O}(N_im_s)$ & $\mathcal{O}(N_iB+Q)$\\
CompCGA & $\mathcal{O}(\frac{m_s N_i}{b})$& $\mathcal{O}(N_im_s)$  & $\mathcal{O}(N_iB+Q)$\\
\textit{NGC} $\alpha=0$  & 0& 0 &$\mathcal{O}(N_iB+m_sN_i)$ \\
\textit{NGC}  & $\mathcal{O}(m_s N_i)$&0 &$\mathcal{O}(N_iB+m_sN_i)$ \\
\textit{CompNGC}  & $\mathcal{O}(\frac{m_s N_i}{b})$ & $\mathcal{O}(N_im_s)$& $\mathcal{O}(N_iB+m_sN_i)$\\
\hline
\end{tabular}
}
\end{center}
\end{table}

\section{Conclusion}
Enabling decentralized training over non-IID data is key for ML applications to efficiently leverage the humongous amounts of user-generated private data.
In this paper, we propose the \textit{Neighborhood Gradient Clustering} (\textit{NGC}) algorithm that improves decentralized learning over non-IID data distributions. Further, we present a compressed version of our algorithm (\textit{CompNGC}) to reduce the communication overhead associated with \textit{NGC}. We theoretically analyze the convergence characteristic and empirically validate the performance of the proposed techniques over different model architectures, graph sizes, and topologies. Finally, we compare the proposed algorithms with the current state-of-the-art decentralized learning algorithm over non-IID data and show superior performance with significantly less compute and memory requirements setting the new state-of-the-art for decentralized learning over non-IID data. 

\section*{Author Contribution:}
Sai Aparna Aketi and Sangamesh Kodge, both worked on developing the algorithm and performed the simulations on computer vision tasks. The simulations for NLP tasks were conducted by Sangamesh Kodge. The theory for the convergence analysis and consensus error bounds was developed by Sai Aparna Aketi. All the authors contributed equally in writing and proofreading the paper.



\bibliography{example_paper}

\begin{thebibliography}{33}
\providecommand{\natexlab}[1]{#1}
\providecommand{\url}[1]{\texttt{#1}}
\expandafter\ifx\csname urlstyle\endcsname\relax
  \providecommand{\doi}[1]{doi: #1}\else
  \providecommand{\doi}{doi: \begingroup \urlstyle{rm}\Url}\fi

\bibitem[Agarwal \& Duchi(2011)Agarwal and Duchi]{agarwal2011distributed}
Agarwal, A. and Duchi, J.~C.
\newblock Distributed delayed stochastic optimization.
\newblock \emph{Advances in neural information processing systems}, 24, 2011.

\bibitem[Aketi et~al.(2021)Aketi, Singh, and Rabaey]{sparsepush}
Aketi, S.~A., Singh, A., and Rabaey, J.
\newblock Sparse-push: Communication-\& energy-efficient decentralized
  distributed learning over directed \& time-varying graphs with non-iid
  datasets.
\newblock \emph{arXiv preprint arXiv:2102.05715}, 2021.

\bibitem[Aketi et~al.(2022)Aketi, Kodge, and Roy]{lp}
Aketi, S.~A., Kodge, S., and Roy, K.
\newblock Low precision decentralized distributed training over iid and non-iid
  data.
\newblock \emph{Neural Networks}, 2022.
\newblock ISSN 0893-6080.
\newblock \doi{https://doi.org/10.1016/j.neunet.2022.08.032}.

\bibitem[Assran et~al.(2019)Assran, Loizou, Ballas, and Rabbat]{sgp}
Assran, M., Loizou, N., Ballas, N., and Rabbat, M.
\newblock Stochastic gradient push for distributed deep learning.
\newblock In \emph{International Conference on Machine Learning}, pp.\
  344--353. PMLR, 2019.

\bibitem[Balu et~al.(2021)Balu, Jiang, Tan, Hedge, Lee, and
  Sarkar]{balu2021decentralized}
Balu, A., Jiang, Z., Tan, S.~Y., Hedge, C., Lee, Y.~M., and Sarkar, S.
\newblock Decentralized deep learning using momentum-accelerated consensus.
\newblock In \emph{ICASSP 2021-2021 IEEE International Conference on Acoustics,
  Speech and Signal Processing (ICASSP)}, pp.\  3675--3679. IEEE, 2021.

\bibitem[Bottou(2010)]{sgd}
Bottou, L.
\newblock Large-scale machine learning with stochastic gradient descent.
\newblock In \emph{Proceedings of COMPSTAT'2010}, pp.\  177--186. Springer,
  2010.

\bibitem[Dean et~al.(2012)Dean, Corrado, Monga, Chen, Devin, Mao, Ranzato,
  Senior, Tucker, Yang, et~al.]{dean2012large}
Dean, J., Corrado, G., Monga, R., Chen, K., Devin, M., Mao, M., Ranzato, M.,
  Senior, A., Tucker, P., Yang, K., et~al.
\newblock Large scale distributed deep networks.
\newblock \emph{Advances in neural information processing systems}, 25, 2012.

\bibitem[Devlin et~al.(2018)Devlin, Chang, Lee, and Toutanova]{bert}
Devlin, J., Chang, M.-W., Lee, K., and Toutanova, K.
\newblock Bert: Pre-training of deep bidirectional transformers for language
  understanding.
\newblock \emph{arXiv preprint arXiv:1810.04805}, 2018.

\bibitem[Esfandiari et~al.(2021)Esfandiari, Tan, Jiang, Balu, Herron, Hegde,
  and Sarkar]{cga}
Esfandiari, Y., Tan, S.~Y., Jiang, Z., Balu, A., Herron, E., Hegde, C., and
  Sarkar, S.
\newblock Cross-gradient aggregation for decentralized learning from non-iid
  data.
\newblock In \emph{International Conference on Machine Learning}, pp.\
  3036--3046. PMLR, 2021.

\bibitem[Goldfarb \& Idnani(1983)Goldfarb and Idnani]{qp}
Goldfarb, D. and Idnani, A.
\newblock A numerically stable dual method for solving strictly convex
  quadratic programs.
\newblock \emph{Mathematical programming}, 27\penalty0 (1):\penalty0 1--33,
  1983.

\bibitem[Haghighat et~al.(2020)Haghighat, Ravichandra-Mouli, Chakraborty,
  Esfandiari, Arabi, and Sharma]{haghighat2020applications}
Haghighat, A.~K., Ravichandra-Mouli, V., Chakraborty, P., Esfandiari, Y.,
  Arabi, S., and Sharma, A.
\newblock Applications of deep learning in intelligent transportation systems.
\newblock \emph{Journal of Big Data Analytics in Transportation}, 2\penalty0
  (2):\penalty0 115--145, 2020.

\bibitem[He et~al.(2016)He, Zhang, Ren, and Sun]{resnet}
He, K., Zhang, X., Ren, S., and Sun, J.
\newblock Deep residual learning for image recognition.
\newblock In \emph{Proceedings of the IEEE conference on computer vision and
  pattern recognition}, pp.\  770--778, 2016.

\bibitem[Hsieh et~al.(2020)Hsieh, Phanishayee, Mutlu, and Gibbons]{skewscout}
Hsieh, K., Phanishayee, A., Mutlu, O., and Gibbons, P.
\newblock The non-{IID} data quagmire of decentralized machine learning.
\newblock In \emph{Proceedings of the 37th International Conference on Machine
  Learning}, volume 119 of \emph{Proceedings of Machine Learning Research},
  pp.\  4387--4398. PMLR, 13--18 Jul 2020.

\bibitem[Husain(2018)]{imagenette}
Husain, H.
\newblock Imagenette - a subset of 10 easily classified classes from the
  imagenet dataset.
\newblock \emph{https://github.com/fastai/imagenette}, 2018.

\bibitem[Karimireddy et~al.(2019)Karimireddy, Rebjock, Stich, and
  Jaggi]{karimireddy2019error}
Karimireddy, S.~P., Rebjock, Q., Stich, S., and Jaggi, M.
\newblock Error feedback fixes signsgd and other gradient compression schemes.
\newblock In \emph{International Conference on Machine Learning}, pp.\
  3252--3261. PMLR, 2019.

\bibitem[Koloskova et~al.(2019)Koloskova, Lin, Stich, and Jaggi]{choco-sgd}
Koloskova, A., Lin, T., Stich, S.~U., and Jaggi, M.
\newblock Decentralized deep learning with arbitrary communication compression.
\newblock \emph{arXiv preprint arXiv:1907.09356}, 2019.

\bibitem[Koloskova et~al.(2020)Koloskova, Loizou, Boreiri, Jaggi, and
  Stich]{koloskova2020unified}
Koloskova, A., Loizou, N., Boreiri, S., Jaggi, M., and Stich, S.
\newblock A unified theory of decentralized sgd with changing topology and
  local updates.
\newblock In \emph{International Conference on Machine Learning}, pp.\
  5381--5393. PMLR, 2020.

\bibitem[Kone{\v{c}}n{\`y} et~al.(2016)Kone{\v{c}}n{\`y}, McMahan, Ramage, and
  Richt{\'a}rik]{federated}
Kone{\v{c}}n{\`y}, J., McMahan, H.~B., Ramage, D., and Richt{\'a}rik, P.
\newblock Federated optimization: Distributed machine learning for on-device
  intelligence.
\newblock 2016.

\bibitem[Krizhevsky et~al.(2014)Krizhevsky, Nair, and Hinton]{cifar}
Krizhevsky, A., Nair, V., and Hinton, G.
\newblock Cifar (canadian institute for advanced research).
\newblock \emph{http://www.cs.toronto.edu/~kriz/cifar.html}, 2014.

\bibitem[LeCun et~al.(1998)LeCun, Bottou, Bengio, and Haffner]{lenet}
LeCun, Y., Bottou, L., Bengio, Y., and Haffner, P.
\newblock Gradient-based learning applied to document recognition.
\newblock \emph{Proceedings of the IEEE}, 86\penalty0 (11):\penalty0
  2278--2324, 1998.

\bibitem[Lian et~al.(2017)Lian, Zhang, Zhang, Hsieh, Zhang, and Liu]{d-psgd}
Lian, X., Zhang, C., Zhang, H., Hsieh, C.-J., Zhang, W., and Liu, J.
\newblock Can decentralized algorithms outperform centralized algorithms? a
  case study for decentralized parallel stochastic gradient descent.
\newblock \emph{Advances in Neural Information Processing Systems}, 30, 2017.

\bibitem[Lin et~al.(2021)Lin, Karimireddy, Stich, and Jaggi]{qgm}
Lin, T., Karimireddy, S.~P., Stich, S., and Jaggi, M.
\newblock Quasi-global momentum: Accelerating decentralized deep learning on
  heterogeneous data.
\newblock In \emph{Proceedings of the 38th International Conference on Machine
  Learning}, volume 139 of \emph{Proceedings of Machine Learning Research},
  pp.\  6654--6665. PMLR, 18--24 Jul 2021.

\bibitem[Liu et~al.(2020)Liu, Brock, Simonyan, and Le]{evonorm}
Liu, H., Brock, A., Simonyan, K., and Le, Q.
\newblock Evolving normalization-activation layers.
\newblock In Larochelle, H., Ranzato, M., Hadsell, R., Balcan, M.~F., and Lin,
  H. (eds.), \emph{Advances in Neural Information Processing Systems},
  volume~33, pp.\  13539--13550. Curran Associates, Inc., 2020.

\bibitem[Nadiradze et~al.(2019)Nadiradze, Sabour, Alistarh, Sharma, Markov, and
  Aksenov]{swarmsgd}
Nadiradze, G., Sabour, A., Alistarh, D., Sharma, A., Markov, I., and Aksenov,
  V.
\newblock Swarmsgd: Scalable decentralized sgd with local updates.
\newblock \emph{arXiv preprint arXiv:1910.12308}, 2019.

\bibitem[Sandler et~al.(2018)Sandler, Howard, Zhu, Zhmoginov, and
  Chen]{mobilnetv2}
Sandler, M., Howard, A., Zhu, M., Zhmoginov, A., and Chen, L.-C.
\newblock Mobilenetv2: Inverted residuals and linear bottlenecks.
\newblock In \emph{Proceedings of the IEEE conference on computer vision and
  pattern recognition}, pp.\  4510--4520, 2018.

\bibitem[Sanh et~al.(2019)Sanh, Debut, Chaumond, and Wolf]{distilbert}
Sanh, V., Debut, L., Chaumond, J., and Wolf, T.
\newblock Distilbert, a distilled version of bert: smaller, faster, cheaper and
  lighter.
\newblock \emph{arXiv preprint arXiv:1910.01108}, 2019.

\bibitem[Simonyan \& Zisserman(2014)Simonyan and Zisserman]{vgg}
Simonyan, K. and Zisserman, A.
\newblock Very deep convolutional networks for large-scale image recognition.
\newblock \emph{arXiv preprint arXiv:1409.1556}, 2014.

\bibitem[Tang et~al.(2018)Tang, Lian, Yan, Zhang, and Liu]{d2}
Tang, H., Lian, X., Yan, M., Zhang, C., and Liu, J.
\newblock $d^2$: Decentralized training over decentralized data.
\newblock In \emph{International Conference on Machine Learning}, pp.\
  4848--4856. PMLR, 2018.

\bibitem[Tang et~al.(2019)Tang, Lian, Qiu, Yuan, Zhang, Zhang, and
  Liu]{deepsqueeze}
Tang, H., Lian, X., Qiu, S., Yuan, L., Zhang, C., Zhang, T., and Liu, J.
\newblock Deepsqueeze: Decentralization meets error-compensated compression.
\newblock \emph{arXiv preprint arXiv:1907.07346}, 2019.

\bibitem[Xiao et~al.(2017)Xiao, Rasul, and Vollgraf]{fmnist}
Xiao, H., Rasul, K., and Vollgraf, R.
\newblock Fashion-mnist: a novel image dataset for benchmarking machine
  learning algorithms.
\newblock \emph{arXiv preprint arXiv:1708.07747}, 2017.

\bibitem[Xiao \& Boyd(2004)Xiao and Boyd]{gossip}
Xiao, L. and Boyd, S.
\newblock Fast linear iterations for distributed averaging.
\newblock \emph{Systems \& Control Letters}, 53\penalty0 (1):\penalty0 65--78,
  2004.

\bibitem[Yu et~al.(2019)Yu, Jin, and Yang]{yu2019linear}
Yu, H., Jin, R., and Yang, S.
\newblock On the linear speedup analysis of communication efficient momentum
  sgd for distributed non-convex optimization.
\newblock \emph{arXiv preprint arXiv:1905.03817}, 2019.

\bibitem[Zhang et~al.(2015)Zhang, Zhao, and LeCun]{agnews}
Zhang, X., Zhao, J., and LeCun, Y.
\newblock Character-level convolutional networks for text classification.
\newblock \emph{Advances in neural information processing systems}, 28, 2015.

\end{thebibliography}
\bibliographystyle{icml2023}

\newpage
\appendix
\onecolumn
\section{Appendix}

\subsection{Proof of Lemma.~\ref{lemma1}}\label{apx:lemma1}
This section presents detailed proof for Lemma.~\ref{lemma1}.
The \textit{NGC} algorithm modifies the local gradients as follows
\begin{equation*}
\begin{split}
     \widetilde{g}^i &= (1-\alpha)\sum_{j\in \mathcal{N}(i)}w_{ji}g^{ji}+ \alpha \sum_{j\in \mathcal{N}(i)}w_{ij}g^{ij}
     =g^{i} + \sum_{j\in \mathcal{N}(i)}w_{ij}(g^{ij}-g^{i}) \hspace{2mm} (\text{assume }\alpha=1)\\
     \widetilde{g}^i & - g^i = \sum_{j\in \mathcal{N}(i)}w_{ij}(\nabla F_j(x,d^j) - \nabla F_i(x,d^i))
\end{split}
\end{equation*}

Now we prove the lemma.~\ref{lemma1} by applying expectation to the above inequality
\begin{equation*}
\begin{split}
\mathbb{E}\bigg[\bigg\|\frac{1}{N}\sum_{i=1}^N(\tilde{\mathbf{g}}^i -\mathbf{g}^i)\bigg\|^2\bigg]
& = \mathbb{E}\bigg[\bigg\|\frac{1}{N}\sum_{i=1}^N \sum_{j\in \mathcal{N}(i)}w_{ij} (\nabla F_j(x,d^j) - \nabla F_i(x,d^i))\bigg\|^2\bigg] \\
& \overset{(a)}{=} \mathbb{E}\bigg[\bigg\|\frac{1}{N} \sum_{i=1}^N \sum_{j\in \mathcal{N}(i)}w_{ij}(\nabla F_j(x,d^j) - \nabla F_i(x,d^i)-\mathbb{E}[\nabla F_j(x,d^j) - \nabla F_i(x,d^i)])\bigg\|^2\bigg] \\
&+ \mathbb{E}\bigg[\bigg\|\frac{1}{N}\sum_{i=1}^N \sum_{j\in \mathcal{N}(i)}w_{ij}(\mathbb{E}[\nabla F_j(x,d^j) - \nabla F_i(x,d^i)])\bigg\|^2\bigg] \\
& \overset{(b)} = \mathbb{E}\bigg[\bigg\|\frac{1}{N}\sum_{i=1}^N \sum_{j\in \mathcal{N}(i)}w_{ij}(\nabla F_j(x,d^j)-\nabla f_j(x)) - (\nabla F_i(x,d^i)-\nabla f_i(x))\bigg\|^2\bigg] \\
&+ \mathbb{E}\bigg[\bigg\|\frac{1}{N}\sum_{i=1}^N \sum_{j\in \mathcal{N}(i)}w_{ij}(\nabla f_j(x)   - \nabla f_i(x))\bigg\|^2\bigg] \\
& \overset{(c)} = \frac{1}{N^2} \sum_{i=1}^N\mathbb{E}\bigg[\bigg\| \sum_{j\in \mathcal{N}(i)}w_{ij}(\nabla F_j(x,d^j)-\nabla f_j(x)) - (\nabla F_i(x,d^i)-\nabla f_i(x))\bigg\|^2\bigg] \\
&+ \mathbb{E}\bigg[\bigg\|\frac{1}{N}\sum_{i=1}^N \sum_{j\in \mathcal{N}(i)}w_{ij}(\nabla f_j(x)   - \nabla f_i(x))\bigg\|^2\bigg] \\
& \overset{(d)} \leq \frac{1}{N^2} \sum_{i=1}^N \sum_{j\in \mathcal{N}(i)}w_{ij}\mathbb{E}\bigg[\bigg\| \nabla F_j(x,d^j)-\nabla f_j(x) - (\nabla F_i(x,d^i)-\nabla f_i(x))\bigg\|^2\bigg] \\
&+ \frac{1}{N}\sum_{i=1}^N \sum_{j\in \mathcal{N}(i)}w_{ij}\mathbb{E}\bigg[\bigg\|\nabla f_j(x)  - \nabla \mathcal{F}(x)+ \nabla \mathcal{F}(x)  - \nabla f_i(x)\bigg\|^2\bigg] \\
& \overset{(e)} \leq \frac{1}{N^2} \sum_{i=1}^N \sum_{j\in \mathcal{N}(i)}w_{ij} 4 \sigma^2 + \frac{1}{N}\sum_{i=1}^N \sum_{j\in \mathcal{N}(i)}w_{ij}\mathbb{E}\bigg[\bigg\|\nabla f_j(x)  - \nabla \mathcal{F}(x)+ \nabla \mathcal{F}(x)  - \nabla f_i(x)\bigg\|^2\bigg] \\
& = \frac{4N \sigma^2}{N^2}  + \frac{1}{N}\sum_{i=1}^N \sum_{j\in \mathcal{N}(i)}w_{ij}\mathbb{E}\bigg[\bigg\|\nabla f_j(x)  - \nabla \mathcal{F}(x)+ \nabla \mathcal{F}(x)  - \nabla f_i(x)\bigg\|^2\bigg] \hspace{4mm} (\because\sum_{j\in \mathcal{N}(i)}w_{ij} =1 )\\
& \overset{(f)}\leq \frac{4 \sigma^2}{N}  + \frac{1}{N}\sum_{i=1}^N \sum_{j\in \mathcal{N}(i)}w_{ij}\bigg(2\mathbb{E}\bigg[\bigg\|\nabla f_j(x)  - \nabla \mathcal{F}(x)\bigg\|^2\bigg] + 2\mathbb{E}\bigg[\bigg\| \nabla f_i(x) - \nabla \mathcal{F}(x) \bigg\|^2\bigg]\bigg) \\
& \overset{(g)} \leq \frac{4 \sigma^2}{N} + \frac{1}{N} \sum_{i=1}^N \sum_{j\in \mathcal{N}(i)}w_{ij} 4 \zeta^2  \\
& = \frac{4 \sigma^2}{N} + 4 \zeta^2
\end{split}
\end{equation*}

(a) identity  $\mathbb{E}[||Z||^2] =  \mathbb{E}[||Z-E[Z]||^2] + \mathbb{E}[||\mathbb{E}[Z]||^2]$ holds for any random vector Z; (b) the fact  $\mathbb{E}[\nabla F_i(x,d^i)] = \nabla f_i(x) $; 
(c) $\nabla F_i(x,d^i)-\nabla f_i(x)$ are independent random vectors with 0 mean and $\mathbb{E}[||\sum_{i=1}^N Z_i||] = \sum_{i=1}^N \mathbb{E}[||Z_i||^2]$ holds when $Z_i$ are independent with mean zero;
(d) jensen's inequality; 
(e) follows from assumption 2  $\mathbb{E}_{ d \sim D_i} || \nabla F_i(x; d) - \nabla f_i(x)||^2 \leq \sigma^2 \hspace{2mm} \forall i \in [1,N]$;
(f) the basic inequality $||a_1 + a_2||^2 \leq 2||a_1||^2 + 2||a_2||^2$ for any vectors $a_1$, $a_2$; and
(g)  follows from assumption 2 that $|| \nabla f_i(x) - \nabla \mathcal{F}(x)||^2 \leq \zeta^2 \hspace{2mm} \forall i \in [1,N]$ \\

$\therefore$ we have following bound given by lemma.~\ref{lemma1}: $ \mathbb{E}[||\widetilde{g}^i  - g^i||^2]\leq 4 (\frac{\sigma^2}{N} +  \zeta^2)$

\subsection{Convergence Analysis}
In this section, we present the proof for our main theorem.~\ref{theorem_1} indicating the convergence of the proposed \textit{NGC} algorithm. Note that without loss of generality, we assume the \textit{NGC mixing weight $\alpha$} to be 1 for the proof. Also, empirically best results for \textit{NGC} are obtained when $\alpha$ is set to 1. Before we proceed to the proof of the theorem, we present a lemma showing that \textit{NGC} achieves consensus among the different agents.

\subsubsection{Bound for Consensus Error} 
\begin{lemma}\label{lemma_2}
Given assumptions 1-3 and lemma~\ref{lemma1}, the distance between the average sequence iterate $\Bar{x}^k$ and the sequence iterates $x_i^k$'s  generated \textit{NGC} (i.e., the consensus error of the proposed algorithm) is given by
\begin{equation}
\label{eq:th1}
    \begin{split}
    \sum_{k=0}^K \frac{1}{N} \sum_{i=1}^N \mathbb{E}[||\Bar{x}_k - x^i_k||^2] \leq & 
    \hspace{4mm}  \frac{14\eta^2 (\sigma^2+\zeta^2)}{(1-\beta)^2 (1-\sqrt{\rho})^2}K+ \frac{6 \eta^2}{(1-\beta)^2(1-\sqrt{\rho})}\sum_{k=0}^{K-1}\mathbb{E}[\|\frac{1}{N}\sum_{i=1}^N\nabla f_i(\mathbf{x}^i_k)\|^2]
    \end{split}
\end{equation}
where  $K \geq 1$ and $\beta \in [0,1)$ is the momentum coefficient. 
\end{lemma}

To prove Lemma~\ref{lemma_2} and Theorem.~\ref{theorem_1}, we follow the similar approach as \cite{cga}. Hence, we also define the following auxiliary sequence along with Lemma~~\ref{lemma_3}
\begin{equation}\label{z_seq}
    \bar{\mathbf{z}}_k:=\frac{1}{1-\beta}\bar{\mathbf{x}}_k-\frac{\beta}{1-\beta}\bar{\mathbf{x}}_{k-1}
\end{equation}
Where $k>0$ and $\mathbf{x}_k$ is obtained by multiplying the update law by $\frac{1}{N}\mathbf{1}\mathbf{1}^\top$, ($\mathbf{1}$ is the column vector with entries being 1). 
\begin{equation}\label{ave_update}
\begin{split}
    &\bar{\mathbf{v}}_k=\beta\bar{\mathbf{v}}_{k-1}-\eta\frac{1}{N}\sum_{i=1}^N\tilde{\mathbf{g}}^i_{k-1}\\
    &\bar{\mathbf{x}}_k=\bar{\mathbf{x}}_{k-1}+\bar{\mathbf{v}}_{k}
\end{split}
\end{equation}
If $k=0$ then $\bar{\mathbf{z}}^k=\bar{\mathbf{x}}^k$. For the rest of the analysis, the initial value will be directly set to $0$.


To prove Lemma.~\ref{lemma_2}, we use the following facts.
\begin{equation}
\label{fact_1}
    \bar{\mathbf{z}}_{k+1}-\bar{\mathbf{z}}_k=-\frac{\eta}{1-\beta}\frac{1}{N}\sum_{i=1}^N\tilde{\mathbf{g}}^i_k.
\end{equation}

\begin{equation}
\label{fact_2}
    \sum_{k=0}^{K-1}\|\bar{\mathbf{z}}_k-\bar{\mathbf{x}}_k\|^2\leq \frac{\eta^2\beta^2}{(1-\beta)^4}\sum_{k=0}^{K-1}\bigg\|\frac{1}{N}\sum_{i=1}^N\tilde{\mathbf{g}}^i_k\bigg\|^2.
\end{equation}
Note that the proof for Eq.~\ref{fact_1} and Eq.~\ref{fact_2} can be found in \cite{cga} as lemma-3 and lemma-4 respectively.

Before proceeding to prove Lemma~\ref{lemma_2}, we introduce some key notations and facts that serve to characterize the Lemma similar to \cite{cga}. 

We define the following notations:
\begin{equation}\label{notation}
    \begin{split}
        &\tilde{\mathbf{G}}_k\triangleq [\tilde{\mathbf{g}}^1_k,\tilde{\mathbf{g}}^2_k,...,\tilde{\mathbf{g}}^N_k]\\
        &\mathbf{V}_k\triangleq[\mathbf{v}^1_k,\mathbf{v}^2_k,...,\mathbf{v}^N_k]\\
        &\mathbf{X}_k\triangleq[\mathbf{x}^1_k,\mathbf{x}^2_k,...,\mathbf{x}^N_k]\\
        &\mathbf{G}_k\triangleq[\mathbf{g}^1_k,\mathbf{g}^2_k,...,\mathbf{g}^N_k]\\
        &\mathbf{H}_k\triangleq[\nabla f_1(\mathbf{x}^1_k),\nabla f_2(\mathbf{x}^2_k),...,\nabla f_N(\mathbf{x}^N_k)]\\
    \end{split}
\end{equation}

We can observe that the above matrices are all with dimension $d\times N$ such that any matrix $\mathbf{A}$ satisfies $\|\mathbf{A}\|_\mathfrak{F}^2=\sum_{i=1}^N\|\mathbf{a}_i\|^2$, where $\mathbf{a}_i$ is the $i$-th column of the matrix $\mathbf{A}$.  Thus, we can obtain that:
\begin{equation}
    \|\mathbf{X}_k(\mathbf{I}-\mathbf{Q})\|_\mathfrak{F}^2=\sum_{i=1}^N\|\mathbf{x}^i_k-\bar{\mathbf{x}}_k\|^2.
\end{equation}

Define $\mathbf{Q}=\frac{1}{N}\mathbf{1}\mathbf{1}^\top$. For each doubly stochastic matrix $\mathbf{W}$, the following properties can be obtained
\begin{equation}\label{fact_3}
\begin{split}
    &\hspace{1mm}(i)\hspace{4.2mm} \mathbf{Q}\mathbf{W}=\mathbf{W}\mathbf{Q};\\
    &\hspace{0.2mm}(ii)\hspace{4mm}(\mathbf{I}-\mathbf{Q})\mathbf{W}=\mathbf{W}(\mathbf{I}-\mathbf{Q});\\
    &\hspace{-.5mm}(iii)\hspace{4.2mm}\text{For any integer $k\geq 1$, $\|(\mathbf{I}-\mathbf{Q})\mathbf{W}\|_\mathfrak{S}\leq(\sqrt{\rho})^k$, where $\|\cdot\|_\mathfrak{S}$ is the spectrum norm of a matrix.}
\end{split}
\end{equation}

Let $\mathbf{A}_i, i\in\{1,2,...,N\}$ be $N$ arbitrary real square matrices. It follows that
\begin{equation}\label{fact_4}
    \|\sum_{i=1}^N\mathbf{A}_i\|^2_\mathfrak{F}\leq \sum_{i=1}^N\sum_{j=1}^N\|\mathbf{A}_i\|_\mathfrak{F}\|\mathbf{A}_j\|_\mathfrak{F}.
\end{equation}

We present the following auxiliary lemmas to prove the lemma.~\ref{lemma_2}
\begin{lemma}\label{lemma_3}
Let all assumptions hold. Let $g^i$ be the unbiased estimate of $\nabla f_i(\mathbf{x}^i)$ at the point $\mathbf{x}^i$ such that $\mathbb{E}[\mathbf{g}^i]=\nabla f_i(\mathbf{x}^i)$, for all $i\in[N]:=\{1,2,...,N\}$. Thus the following relationship holds
\begin{equation}
    \mathbb{E}\bigg[\bigg\|\frac{1}{N}\sum_{i=1}^N\tilde{\mathbf{g}}^i\bigg\|^2\bigg]\leq  \frac{10\sigma^2}{N} + 8\zeta^2 + 2\mathbb{E}\bigg[\bigg\|\frac{1}{N}\sum_{i=1}^N\nabla f_i(\mathbf{x}^i)\bigg\|^2\bigg]
\end{equation}
\end{lemma}

\textit{Proof for Lemma~\ref{lemma_3}:}
\begin{equation*}
\begin{split}
    \mathbb{E}\bigg[\bigg\|\frac{1}{N}\sum_{i=1}^N\tilde{\mathbf{g}}^i\bigg\|^2\bigg] &=
    \mathbb{E}\bigg[\bigg\|\frac{1}{N}\sum_{i=1}^N(\tilde{\mathbf{g}}^i -\mathbf{g}^i+\mathbf{g}^i )\bigg\|^2\bigg] \\
    &= \mathbb{E}\bigg[\bigg\|\frac{1}{N}\sum_{i=1}^N(\tilde{\mathbf{g}}^i -\mathbf{g}^i)+\frac{1}{N}\sum_{i=1}^N \mathbf{g}^i \bigg\|^2\bigg]\\
    &\overset{a}{\leq} 2\mathbb{E}\bigg[\bigg\|\frac{1}{N}\sum_{i=1}^N(\tilde{\mathbf{g}}^i -\mathbf{g}^i)\bigg\|^2\bigg]+2\mathbb{E}\bigg[\bigg\|\frac{1}{N}\sum_{i=1}^N \mathbf{g}^i \bigg\|^2\bigg]\\
    & \overset{b}\leq 8(\frac{\sigma^2}{N} + \zeta^2) + \frac{2\sigma^2}{N} + 2\mathbb{E}\bigg[\bigg\|\frac{1}{N}\sum_{i=1}^N\nabla f_i(\mathbf{x}^i)\bigg\|^2\bigg] \\
    & = \frac{10\sigma^2}{N} + 8\zeta^2 + 2\mathbb{E}\bigg[\bigg\|\frac{1}{N}\sum_{i=1}^N\nabla f_i(\mathbf{x}^i)\bigg\|^2\bigg] \\
\end{split}
\end{equation*}

(a) refers to the fact that the inequality $\|\mathbf{a}+\mathbf{b}\|^2\leq 2\|\mathbf{a}\|^2+2\|\mathbf{b}\|^2$. (b) The first term uses Lemma~\ref{lemma1} and the second term uses the conclusion of Lemma $1$ in ~\cite{yu2019linear}.

\begin{lemma}\label{lemma_3a}
Let all assumptions hold. Let $g^i$ be the unbiased estimate of $\nabla f_i(\mathbf{x}^i)$ at the point $\mathbf{x}^i$ such that $\mathbb{E}[\mathbf{g}^i]=\nabla f_i(\mathbf{x}^i)$, for all $i\in[N]:=\{1,2,...,N\}$. Thus the following relationship holds
\begin{equation}
    \mathbb{E}\bigg[\bigg\|\tilde{\mathbf{G}}_{\tau}-\mathbf{G}_{\tau}\bigg\|^2_\mathfrak{F}\bigg] \leq  4N (\sigma^2+\zeta^2)
\end{equation}
\end{lemma}

\textit{Proof for Lemma~\ref{lemma_3a} is similar to Lemma~\ref{lemma1} and is as follows:}
\begin{equation*}
\begin{split}
    \mathbb{E}\bigg[\bigg\|\tilde{\mathbf{G}}_{\tau}-\mathbf{G}_{\tau}\bigg\|^2_\mathfrak{F}\bigg] &=
    \sum_{i=1}^N\mathbb{E}\bigg[\bigg\| \sum_{j\in \mathcal{N}(i)}w_{ij} (\nabla F_j(x,d^j) - \nabla F_i(x,d^i))\bigg\|^2\bigg] \\
    &  = \sum_{i=1}^N \mathbb{E}\bigg[\bigg\| \sum_{j\in \mathcal{N}(i)}w_{ij}(\nabla F_j(x,d^j)-\nabla f_j(x)) - (\nabla F_i(x,d^i)-\nabla f_i(x))\bigg\|^2\bigg] \\
    &+ \sum_{i=1}^N \mathbb{E}\bigg[\bigg\| \sum_{j\in \mathcal{N}(i)}w_{ij}(\nabla f_j(x)   - \nabla f_i(x))\bigg\|^2\bigg] \\
    & = \mathbb{E}\bigg[\bigg\|\sum_{j\in \mathcal{N}(i)}w_{ij}(\nabla F_j(x,d^j) - \nabla F_i(x,d^i)-\mathbb{E}[\nabla F_j(x,d^j) - \nabla F_i(x,d^i)])\bigg\|^2\bigg] \\
    & \leq  \sum_{i=1}^N \sum_{j\in \mathcal{N}(i)}w_{ij}\mathbb{E}\bigg[\bigg\| \nabla F_j(x,d^j)-\nabla f_j(x) - (\nabla F_i(x,d^i)-\nabla f_i(x))\bigg\|^2\bigg] \\
    &+ \sum_{i=1}^N \sum_{j\in \mathcal{N}(i)}w_{ij}\mathbb{E}\bigg[\bigg\|\nabla f_j(x)  - \nabla \mathcal{F}(x)+ \nabla \mathcal{F}(x)  - \nabla f_i(x)\bigg\|^2\bigg] \\
    & \leq 4N(\sigma^2 + \zeta^2) \hspace{4mm} (\text{follows from assumption 2) }
\end{split}
\end{equation*}


The properties shown in Eq.~\ref{fact_3} and~\ref{fact_4} have been well established and in this context, we skip the proof here. We are now ready to prove Lemma~\ref{lemma_2}.

\textit{Proof for Lemma~\ref{lemma_2}:}
Since $ \mathbf{X}_{k} =\mathbf{X}_{k-1}\mathbf{W} + \mathbf{V}_k$ we have:

\begin{equation}
\begin{split}
    &\mathbf{X}_{k}(\mathbf{I}-\mathbf{Q}) =\mathbf{X}_{k-1}(\mathbf{I}-\mathbf{Q})\mathbf{W} + \mathbf{V}_k(\mathbf{I}-\mathbf{Q})
\end{split}
\end{equation}

Applying the above equation $k$ times we have: 

\begin{equation}
\begin{split}
    &\mathbf{X}_{k}(\mathbf{I}-\mathbf{Q}) =\mathbf{X}_{0}(\mathbf{I}-\mathbf{Q})\mathbf{W}^k + \sum_{\tau=1}^{k}\mathbf{V}_{\tau}(\mathbf{I}-\mathbf{Q})\mathbf{W}^{k-\tau} \overset{\mathbf{X}_{0}=0}{=}\sum_{\tau=1}^{k}\mathbf{V}_{\tau}(\mathbf{I}-\mathbf{Q})\mathbf{W}^{k-\tau}
\end{split}
\end{equation} 
As $\bar{\mathbf{V}}_k=\beta\bar{\mathbf{V}}_{k-1}-\eta\frac{1}{N}\sum_{i=1}^N\tilde{\mathbf{G}}^i_{k-1} \overset{\mathbf{V}_{0}=0}{=} -\eta\frac{1}{N}\sum_{i=1}^N\tilde{\mathbf{G}}^i_{k-1}$, we can get:

\begin{equation}
\begin{split}
    \mathbf{X}_k (\mathbf{I}-\mathbf{Q}) &= 
     -\eta\sum_{\tau=1}^k\sum_{l=0}^{\tau-1}\tilde{\mathbf{G}}_{l}\beta^{\tau-1-l}(\mathbf{I}-\mathbf{Q}) \mathbf{W}^{k-\tau} \\
     &= -\eta\sum_{\tau=1}^k\sum_{l=0}^{\tau-1}\tilde{\mathbf{G}}_{l}\beta^{\tau-1-l}\mathbf{W}^{k-\tau}(\mathbf{I}-\mathbf{Q}) -\eta\sum_{n=1}^{k-1}\tilde{\mathbf{G}}_{n}[\sum_{l=n+1}^{k}\beta^{l-1-n}\mathbf{W}^{k-1-n}(\mathbf{I}-\mathbf{Q})\\
    & = -\eta\sum_{\tau=0}^{k-1}\frac{1-\beta^{k-\tau}}{1-\beta}\tilde{\mathbf{G}}_{\tau}(\mathbf{I}-\mathbf{Q})\mathbf{W}^{k-1-\tau}.
\end{split}
\end{equation}

Therefore, for $k\geq 1$, we have:

\begin{equation}
    \begin{split}\label{expand}
        \mathbb{E}\bigg[\bigg\|\mathbf{X}_k(\mathbf{I}-\mathbf{Q})\bigg\|^2_\mathfrak{F}\bigg] &= \eta^2 \mathbb{E}\bigg[\bigg\|\sum_{\tau=0}^{k-1}\frac{1-\beta^{k-\tau}}{1-\beta}\tilde{\mathbf{G}}_{\tau}(\mathbf{I}-\mathbf{Q})\mathbf{W}^{k-1-\tau}\bigg\|^2_\mathfrak{F}\bigg]\\
        & \overset{a}{\leq} \underbrace{2\eta^2\mathbb{E}\bigg[\bigg\|\sum_{\tau=0}^{k-1}\frac{1-\beta^{k-\tau}}{1-\beta}(\tilde{\mathbf{G}}_{\tau}-\mathbf{G}_{\tau})(\mathbf{I}-\mathbf{Q})\mathbf{W}^{k-1-\tau}\bigg\|^2_\mathfrak{F}\bigg]}_{I}+ \\
        &\hspace{4mm}\underbrace{2\eta^2\mathbb{E}\bigg[\bigg\|\sum_{\tau=0}^{k-1}\frac{1-\beta^{k-\tau}}{1-\beta}\mathbf{G}_{\tau}(\mathbf{I}-\mathbf{Q})\mathbf{W}^{k-1-\tau}\bigg\|^2_\mathfrak{F}\bigg]}_{II}
    \end{split}
\end{equation}
(a) follows from the inequality $\|\mathbf{A}+\mathbf{B}\|_\mathfrak{F}^2\leq 2\|\mathbf{A}\|_\mathfrak{F}^2+2\|\mathbf{B}\|_\mathfrak{F}^2$.

We develop upper bounds of the term \textbf{I}:
\begin{equation}\label{I}
    \begin{split}
        \mathbb{E}\bigg[\bigg\|\sum_{\tau=0}^{k-1}\frac{1-\beta^{k-\tau}}{1-\beta}(\tilde{\mathbf{G}}_{\tau}-\mathbf{G}_{\tau})(\mathbf{I}-\mathbf{Q})\mathbf{W}^{k-1-\tau}\bigg\|^2_\mathfrak{F}\bigg] & \overset{a}{\leq} \sum_{\tau=0}^{k-1}\mathbb{E}\bigg[\bigg\|\frac{1-\beta^{k-\tau}}{1-\beta}(\tilde{\mathbf{G}}_{\tau}-\mathbf{G}_{\tau})(\mathbf{I}-\mathbf{Q})\mathbf{W}^{k-1-\tau}\bigg\|^2_\mathfrak{F}\bigg]\\
        & \overset{b}{\leq} \frac{1}{(1-\beta)^2}\sum_{\tau=0}^{k-1}\rho^{k-1-\tau}\mathbb{E}\bigg[\bigg\|\tilde{\mathbf{G}}_{\tau}-\mathbf{G}_{\tau}\bigg\|^2_\mathfrak{F}\bigg]\\
        & \overset{c}{\leq} \frac{4}{(1-\beta)^2}\sum_{\tau=0}^{k-1}\rho^{k-1-\tau}N (\sigma^2 + \zeta^2) \overset{d}{\leq} \frac{4N(\sigma^2 + \zeta^2)}{(1-\beta)^2(1-\rho)}
    \end{split}
\end{equation}
(a) follows from Jensen's inequality. (b) follows from the inequality $|\frac{1-\beta^{k-\tau}}{1-\beta}| \leq \frac{1}{1-\beta} $. (c) follows from Lemma~\ref{lemma_2} and Frobenius norm. (d) follows from Assumption 3.

We then proceed to find the upper bound for term \textbf{II}.

\begin{equation}\label{II}
    \begin{split}
        \mathbb{E}\bigg[\bigg\|\sum_{\tau=0}^{k-1}\frac{1-\beta^{k-\tau}}{1-\beta}\mathbf{G}_{\tau}(\mathbf{I}-\mathbf{Q})\mathbf{W}^{k-1-\tau}\bigg\|^2_\mathfrak{F}\bigg] & \overset{a}{\leq} 
        \sum_{\tau=0}^{k-1}\sum_{\tau^\prime=0}^{k-1}\mathbb{E}\bigg[\bigg\|\frac{1-\beta^{k-\tau}}{1-\beta}\mathbf{G}_{\tau}(\mathbf{I}-\mathbf{Q})\mathbf{W}^{k-1-\tau}\bigg\|_\mathfrak{F}\\
        & \hspace{30mm} \bigg\|\frac{1-\beta^{k-\tau}}{1-\beta}\mathbf{G}_{\tau^\prime}(\mathbf{I}-\mathbf{Q})\mathbf{W}^{k-1-\tau^\prime}\bigg\|_\mathfrak{F}\bigg]\\
        & \leq \frac{1}{(1-\beta)^2}\sum_{\tau=0}^{k-1}\sum_{\tau^\prime=0}^{k-1}\rho^{(k-1-\frac{\tau+\tau^\prime}{2})}\mathbb{E}\bigg[\|\mathbf{G}_{\tau}\|_\mathfrak{F}\|\mathbf{G}_{\tau^\prime}\|_\mathfrak{F}\bigg] \\
        &\overset{b}{\leq} \frac{1}{(1-\beta)^2}\sum_{\tau=0}^{k-1}\sum_{\tau^\prime=0}^{k-1}\rho^{(k-1-\frac{\tau+\tau^\prime}{2})}\bigg(\frac{1}{2}\mathbb{E}[\|\mathbf{G}_{\tau}\|_\mathfrak{F}^2]+\frac{1}{2}\mathbb{E}[\|\mathbf{G}_{\tau^\prime}\|_\mathfrak{F}^2]\bigg)\\
        &= \frac{1}{(1-\beta)^2}\sum_{\tau=0}^{k-1}\sum_{\tau^\prime=0}^{k-1}\rho^{(k-1-\frac{\tau+\tau^\prime}{2})}\mathbb{E}[\|\mathbf{G}_{\tau}\|_\mathfrak{F}^2]\\
        &\overset{c}{\leq} \frac{1}{(1-\beta)^2(1-\sqrt{\rho})}\sum_{\tau=0}^{k-1}\rho^{(\frac{k-1-\tau}{2})}\mathbb{E}[\|\mathbf{G}_{\tau}\|_\mathfrak{F}^2]
    \end{split}
\end{equation}

(a) follows from Eq.~\ref{fact_4}. (b) follows from the inequality $xy \leq \frac{1}{2}(x^2+y^2)$ for any two real numbers $x,y$. (c) is derived using $\sum_{\tau_1=0}^{k-1}\rho^{k-1-\frac{\tau_1+\tau}{2}} \leq \frac{\rho^{\frac{k-1-\tau}{2}}}{1-\sqrt{\rho}}$.

We then proceed with finding the bounds for $\mathbb{E}[\|\mathbf{G}_{\tau}\|_\mathfrak{F}^2]$:

\begin{equation}\label{E_G}
    \begin{split}
        \mathbb{E}[\|\mathbf{G}_\tau\|^2_\mathfrak{F}] &= \mathbb{E}[\|\mathbf{G}_\tau- \mathbf{H}_\tau+ \mathbf{H}_\tau-\mathbf{H}_\tau \mathbf{Q}+\mathbf{H}_\tau \mathbf{Q}\|^2_\mathfrak{F}]\\
        & \leq 3\mathbb{E}[\|\mathbf{G}_\tau- \mathbf{H}_\tau\|^2_\mathfrak{F}]+ 3\mathbb{E}[\|\mathbf{H}_\tau(I-\mathbf{Q})\|^2_\mathfrak{F}]+3\mathbb{E}[\|\mathbf{H}_\tau \mathbf{Q}\|^2_\mathfrak{F}]\\
        &\overset{a}{\leq} 3N\sigma^2+3N\zeta^2+3 \mathbb{E}[\|\frac{1}{N}\sum_{i=1}^N\nabla f_i(\mathbf{x}^i_\tau)\|^2]
    \end{split}
\end{equation}

(a) holds because $\mathbb{E}[\|\mathbf{H}_\tau \mathbf{Q}\|^2_\mathfrak{F}]\leq\mathbb{E}[\|\frac{1}{N}\sum_{i=1}^N\nabla f_i(\mathbf{x}^i_\tau)\|^2]$

Substituting (\ref{E_G}) in ~(\ref{II}):

\begin{equation}\label{prefinal}
    \begin{split}
         \mathbb{E}\bigg[\bigg\|\sum_{\tau=0}^{k-1}\frac{1-\beta^{k-\tau}}{1-\beta}\mathbf{G}_{\tau}(\mathbf{I}-\mathbf{Q})\mathbf{W}^{k-1-\tau}\bigg\|^2_\mathfrak{F}\bigg] &\leq \frac{1}{(1-\beta)^2(1-\sqrt{\rho})}\sum_{\tau=0}^{k-1}\rho^{(\frac{k-1-\tau}{2})}\bigg[3N\sigma^2+3N\zeta^2+3N \mathbb{E}[\|\frac{1}{N}\sum_{i=1}^N\nabla f_i(\mathbf{x}^i_\tau)\|^2]\bigg]\\
        &\leq \frac{3N(\sigma^2+\zeta^2)}{(1-\beta)^2(1-\sqrt{\rho})^2}+\frac{3N}{(1-\beta)^2(1-\sqrt{\rho})}\sum_{\tau=0}^{k-1}\rho^{(\frac{k-1-\tau}{2})}\mathbb{E}[\|\frac{1}{N}\sum_{i=1}^N\nabla f_i(\mathbf{x}^i_\tau)\|^2]
    \end{split}
\end{equation}

substituting~(\ref{prefinal}) and (\ref{I}) into the main inequality (\ref{expand}):

\begin{equation}\label{final1}
    \begin{split}
        \mathbb{E}\bigg[\bigg\|\mathbf{X}_k(\mathbf{I}-\mathbf{Q})\bigg\|^2_\mathfrak{F}\bigg] &\leq \frac{8\eta^2N(\sigma^2 + \zeta^2)}{(1-\beta)^2(1-\rho)}+
        \frac{2\eta^2}{(1-\beta)^2(1-\sqrt{\rho})}\bigg(\frac{3N(\sigma^2)}{1-\sqrt{\rho}}+\frac{3N(\zeta^2)}{1-\sqrt{\rho}}+\\
        &\hspace{75mm} 3N\sum_{\tau=0}^{k-1}\rho^{(\frac{k-1-\tau}{2})}\mathbb{E}[\|\frac{1}{N}\sum_{i=1}^N\nabla f_i(\mathbf{x}^i_\tau)\|^2]\bigg)\\
        &= \frac{2\eta^2}{(1-\beta)^2}\bigg(\frac{4N (\sigma^2 + \zeta^2)}{1-\rho}+\frac{3N\sigma^2}{(1-\sqrt{\rho})^2}+\frac{3N\zeta^2}{(1-\sqrt{\rho})^2}\bigg)+ \\
        &\hspace{30mm} \frac{6N\eta^2}{(1-\beta)^2(1-\sqrt{\rho})}\sum_{\tau=0}^{k-1}\rho^{(\frac{k-1-\tau}{2})}\mathbb{E}[\|\frac{1}{N}\sum_{i=1}^N\nabla f_i(\mathbf{x}^i_\tau)\|^2]\\
        &\leq \frac{14N\eta^2}{(1-\beta)^2}\bigg(\frac{\sigma^2}{(1-\sqrt{\rho})^2}+\frac{\zeta^2}{(1-\sqrt{\rho})^2}\bigg)+\frac{6N\eta^2}{(1-\beta)^2(1-\sqrt{\rho})}\sum_{\tau=0}^{k-1}\rho^{(\frac{k-1-\tau}{2})}\mathbb{E}[\|\frac{1}{N}\sum_{i=1}^N\nabla f_i(\mathbf{x}^i_\tau)\|^2]
    \end{split}
\end{equation}

Summing over $k\in\{1,\dots, K-1\}$ and noting that $\mathbb{E}\bigg[\bigg\|\mathbf{X}_0(\mathbf{I}-\mathbf{Q})\bigg\|^2_\mathfrak{F}\bigg] = 0$:

\begin{equation}\label{final2}
    \begin{split}
        \sum_{k=1}^{K-1}\mathbb{E}\bigg[\bigg\|\mathbf{X}_k(\mathbf{I}-\mathbf{Q})\bigg\|^2_\mathfrak{F}\bigg] & \leq CK + \frac{6N\eta^2}{(1-\beta)^2(1-\sqrt{\rho})}\sum_{k=1}^{K-1}\sum_{\tau=0}^{k-1}\rho^{(\frac{k-1-\tau}{2})}\mathbb{E}[\|\frac{1}{N}\sum_{i=1}^N\nabla f_i(\mathbf{x}^i_\tau)\|^2] \\ 
        &\leq CK + \frac{6N\eta^2}{(1-\beta)^2(1-\sqrt{\rho})}\sum_{k=0}^{K-1}\frac{1-\rho^{(\frac{K-1-k}{2})}}{1-\sqrt{\rho}}\mathbb{E}[\|\frac{1}{N}\sum_{i=1}^N\nabla f_i(\mathbf{x}^i_k)\|^2]\\ 
        &\leq CK + \frac{6N\eta^2}{(1-\beta)^2(1-\sqrt{\rho})}\sum_{k=0}^{K-1}\mathbb{E}[\|\frac{1}{N}\sum_{i=1}^N\nabla f_i(\mathbf{x}^i_k)\|^2]
    \end{split}
\end{equation}
Where $C = \frac{14N\eta^2}{(1-\beta)^2}\bigg(\frac{\sigma^2}{(1-\sqrt{\rho})^2}+\frac{\zeta^2}{(1-\sqrt{\rho})^2}\bigg)$.

Dividing both sides by $N$:

\begin{equation}\label{final2-2}
    \begin{split}
        \sum_{k=1}^{K-1}\frac{1}{N}\mathbb{E}\bigg[\bigg\|\mathbf{X}_k(\mathbf{I}-\mathbf{Q})\bigg\|^2_\mathfrak{F}\bigg] \leq &
        \frac{14\eta^2}{(1-\beta)^2}\bigg(\frac{ \sigma^2}{(1-\sqrt{\rho})^2} +\frac{ \zeta^2}{(1-\sqrt{\rho})^2}\bigg)K +\\
        & \frac{6 \eta^2}{(1-\beta)^2(1-\sqrt{\rho})}\sum_{k=0}^{K-1}\mathbb{E}[\|\frac{1}{N}\sum_{i=1}^N\nabla f_i(\mathbf{x}^i_k)\|^2]
    \end{split}
\end{equation}

Hence, we obtain the following as the lemma.~\ref{lemma_2}:

\begin{equation}
    \begin{split}
        \sum_{k=0}^{K-1}\frac{1}{N}\sum_{i=1}^N\mathbb{E}\bigg[\bigg\|\bar{\mathbf{x}}_k-\mathbf{x}^i_k\bigg\|^2\bigg]\leq &
        \frac{14\eta^2 (\sigma^2+\zeta^2)}{(1-\beta)^2 (1-\sqrt{\rho})^2}K+ \frac{6 \eta^2}{(1-\beta)^2(1-\sqrt{\rho})}\sum_{k=0}^{K-1}\mathbb{E}[\|\frac{1}{N}\sum_{i=1}^N\nabla f_i(\mathbf{x}^i_k)\|^2]
    \end{split}
\end{equation}

\subsubsection{Proof for Theorem.~\ref{theorem_1}}\label{apx:theorem_1}

\textit{Proof:}
Using the L-smoothness properties for  $\mathcal{F}$ we have:
\begin{equation}\label{main0}
    \mathbb{E}[\mathcal{F}(\bar{\mathbf{z}}_{k+1})] \leq \mathbb{E}[\mathcal{F}(\bar{\mathbf{z}}_{k})]+\mathbb{E}[\langle\nabla\mathcal{F}(\bar{\mathbf{z}}_{k}),\bar{\mathbf{z}}_{k+1}- \bar{\mathbf{z}}_{k}\rangle]+ \frac{L}{2} \mathbb{E}[\|\bar{\mathbf{z}}_{k+1} - \bar{\mathbf{z}}_{k}\|^2 ]
\end{equation}

Using Eq.~\ref{fact_1} we have:
\begin{equation}
\label{main1}
\begin{split}
&\mathbb{E}[\langle\nabla\mathcal{F}(\bar{\mathbf{z}}_{k}),\bar{\mathbf{z}}_{k+1}- \bar{\mathbf{z}}_{k}\rangle] =
\frac{-\eta}{1-\beta}\mathbb{E}[\langle\nabla\mathcal{F}(\bar{\mathbf{z}}_{k}),\frac{1}{N}\sum_{i=1}^{N}\tilde{\mathbf{g}}^i_{k}\rangle] =\\ 
&\underbrace{\frac{-\eta}{1-\beta}\mathbb{E}[\langle\nabla\mathcal{F}(\bar{\mathbf{z}}_{k})- \nabla\mathcal{F}(\bar{\mathbf{x}}_{k}),\frac{1}{N}\sum_{i=1}^{N}\tilde{\mathbf{g}}^i_{k}\rangle]}_{T_1}-\underbrace{\frac{\eta}{1-\beta}\mathbb{E}[\langle\nabla\mathcal{F}(\bar{\mathbf{x}}_{k}),\frac{1}{N}\sum_{i=1}^{N}\tilde{\mathbf{g}}^i_{k}\rangle]}_{T_2}
\end{split}
\end{equation}

We proceed by bounding $T_1$:

\begin{equation}\label{eq-t1}
\begin{split}
 & \frac{-\eta}{1-\beta}\mathbb{E}[\langle\nabla\mathcal{F}(\bar{\mathbf{z}}_{k})- \nabla\mathcal{F}(\bar{\mathbf{x}}_{k}),\frac{1}{N}\sum_{i=1}^{N}\tilde{\mathbf{g}}^i_{k}\rangle]\\
 & \overset{(i)}{\leq} \frac{(1-\beta)}{2\beta L}\mathbb{E}[\|\nabla\mathcal{F}(\bar{\mathbf{z}}_{k})- \nabla\mathcal{F}(\bar{\mathbf{x}}_{k})\|^2]+ \frac{\beta L \eta^{2}}{2(1-\beta)^{3}}\mathbb{E}[\| \frac{1}{N} \sum_{i=1}^{N} \tilde{\mathbf{g}}_{k}^{i} \|^{2}] \\
 & \overset{(ii)}{\leq} \frac{(1-\beta)L}{2 \beta}\mathbb{E}[\left\|\bar{\mathbf{z}}_{k}-\bar{\mathbf{x}}_{k}\right\|^{2}]+\frac{\beta L \eta^2}{2(1-\beta)^{3}}\mathbb{E}[\|\frac{1}{N} \sum_{i=1}^{N} \tilde{\mathbf{g}}_{k}^{i}\|^{2}]
\end{split}
\end{equation}

(i)  holds as 
$\langle \mathbf{a},\mathbf{b} \rangle \leq \frac{1}{2}\|\mathbf{a}\|^2 + \frac{1}{2}\|\mathbf{b}\|^2$ where 
$\mathbf{a}=\frac{\sqrt{1-\beta}}{\sqrt{\beta L}} (\nabla\mathcal{F}(\bar{\mathbf{z}}_{k})- \nabla\mathcal{F}(\bar{\mathbf{x}}_{k}))$ 
and $\mathbf{b} = -\frac{\eta\sqrt{\beta L}}{(1-\beta)^{\frac{3}{2}}}\frac{1}{N} \sum_{i=1}^{N}\tilde{\mathbf{g}}^i_{k}$.
 and (ii) uses the fact that $\mathcal{F}$ is L-smooth.

We split the term $T_2$ as follows:
\begin{equation}\label{helper-t2}
\begin{split}
 &\langle\nabla \mathcal{F}\left(\bar{\mathbf{x}}_{k}\right), \frac{1}{N} \sum_{i=1}^{N} \tilde{\mathbf{g}}_{k}^{i}\rangle= \langle\nabla \mathcal{F}\left(\bar{\mathbf{x}}_{k}\right), \frac{1}{N} \sum_{i=1}^{N}\left(\tilde{\mathbf{g}}_{k}^{i}-\mathbf{g}_{k}^{i}+\mathbf{g}_{k}^{i}\right)\rangle =\\
&\underbrace{\langle\nabla \mathcal{F}\left(\bar{\mathbf{x}}_{k}\right), \frac{1}{N} \sum_{i=1}^{N}\left(\tilde{\mathbf{g}}_{k}^{i}-\mathbf{g}_{k}^{i}\right)\rangle}_{T_3}+\underbrace{
\langle\nabla \mathcal{F}\left(\bar{\mathbf{x}}_{k}\right) , \frac{1}{N} \sum_{i=1}^{N}{\mathbf{g}}_{k}^{i}\rangle}_{T_4}
\end{split}
\end{equation}

Now, We first analyze the term $T_3$:
\begin{equation}\label{helper-t3}
    \begin{split}
        &\frac{-\eta}{(1-\beta)}\mathbb{E}[\langle\nabla \mathcal{F}\left(\bar{\mathbf{x}}_{k}\right), \frac{1}{N} \sum_{i=1}^{N}\left(\tilde{\mathbf{g}}_{k}^{i}-\mathbf{g}_{k}^{i}\right)\rangle] \\&\leq \frac{(1-\beta)}{2 \beta L}\mathbb{E}[\|\nabla \mathcal{F}(\bar{\mathbf{x}}_{k})\|^{2}]+\frac{\eta^2 \beta L }{2(1-\beta)^{3}}\mathbb{E}[\|\frac{1}{N} \sum_{i=1}^{N} (\tilde{\mathbf{g}}_{k}^{i}- \mathbf{g}_k^i)\|^{2}]
    \end{split}
\end{equation}

This  holds as 
$\langle \mathbf{a},\mathbf{b} \rangle \leq \frac{1}{2}\|\mathbf{a}\|^2 + \frac{1}{2}\|\mathbf{b}\|^2$ where 
$\mathbf{a}=\frac{- \sqrt{1-\beta}}{\sqrt{\beta L}} \nabla \mathcal{F}(\bar{\mathbf{x}}_k)$ and $\mathbf{b} = \frac{\eta\sqrt{ \beta L}}{(1-\beta)^{\frac{3}{2}}}\frac{1}{N} \sum_{i=1}^{N}(\tilde{\mathbf{g}}_{k}^{i}-\mathbf{g}_{k}^{i})$.

Analyzing the term $T_4$:

\begin{equation}
\label{helper2-2}
    \begin{split}
       \mathbb{E}\bigg[\langle\nabla \mathcal{F}\left(\bar{\mathbf{x}}_{k}\right) , \frac{1}{N} \sum_{i=1}^{N}{\mathbf{g}}_{k}^{i}\rangle\bigg] =  \mathbb{E}\bigg[\langle\nabla\mathcal{F}(\bar{\mathbf{x}}_{k}),\frac{1}{N}\sum_{i=1}^{N}\nabla f_i(\mathbf{x}^i_k)\rangle\bigg]
    \end{split}
\end{equation}

Using the equity $\langle \mathbf{a},\mathbf{b} \rangle = \frac{1}{2}[\|\mathbf{a}\|^2 + \|\mathbf{b}\|^2 - \|\mathbf{a}-\mathbf{b}\|^2]$, we have :

\begin{equation}
\label{helper2-2-1}
\begin{split}
\langle\nabla \mathcal{F}\left(\bar{\mathbf{x}}_{k}\right), \frac{1}{N} \sum_{i=1}^{N} \nabla f_{i}\left(\mathbf{x}_{k}^{i}\right)\rangle &=\frac{1}{2}\left(\|\nabla \mathcal{F}\left(\bar{\mathbf{x}}_{k}\right)\|^{2}+\| \frac{1}{N} \sum_{i=1}^{N}
\nabla f_{i}(\mathbf{x}_{k}^{i})\|^{2}-\| \nabla \mathcal{F}(\bar{\mathbf{x}}_{k})-\frac{1}{N} \sum_{i=1}^{N} \nabla f_{i}(\mathbf{x}_{k}^{i}) \|^{2}\right) \\
&\overset{a}{\geq}
\frac{1}{2}\left(\|\nabla \mathcal{F}(\bar{\mathbf{x}}_{k})\|^{2}+
\|\frac{1}{N} \sum_{i=1}^{N} \nabla f_{i}(\mathbf{x}_{k}^{i})\|^{2}-L^{2} \frac{1}{N} \sum_{i=1}^{N}\|\bar{\mathbf{x}}_{k}-\mathbf{x}_{k}^{i}\|^{2}\right)
\end{split}
\end{equation}

(a) holds because $\|\nabla\mathcal{F}(\bar{\mathbf{x}}_{k})- \frac{1}{N}\sum_{i=1}^{N}\nabla f_i(\mathbf{x}^i_k)\|^2 = \|\frac{1}{N}\sum_{i=1}^{N}\nabla f_i(\bar{\mathbf{x}}_{k})- \frac{1}{N}\sum_{i=1}^{N}\nabla f_i(\mathbf{x}^i_k)\|^2 \leq \frac{1}{N}\sum_{i=1}^{N}\|\nabla f_i(\bar{\mathbf{x}}_{k})- \nabla f_i(\mathbf{x}^i_k)\|^2 \leq \frac{1}{N}\sum_{i=1}^{N} L^2 \|\bar{\mathbf{x}}_{k}- \mathbf{x}^i_k\|^2$. \\

Substituting~(\ref{helper2-2-1}) into~(\ref{helper2-2}), we have 
\begin{equation}
\label{helper-t4}
\begin{split}
 \mathbb{E}\bigg[\langle\nabla \mathcal{F}\left(\bar{\mathbf{x}}_{k}\right) , \frac{1}{N} \sum_{i=1}^{N}{\mathbf{g}}_{k}^{i}\rangle\bigg]
&\geq
\frac{1}{2}\left(\mathbb{E}[\|\nabla \mathcal{F}(\bar{\mathbf{x}}_{k})\|^{2}]+
\mathbb{E}[\|\frac{1}{N} \sum_{i=1}^{N} \nabla f_{i}(\mathbf{x}_{k}^{i})\|^{2}]-L^{2} \frac{1}{N} \sum_{i=1}^{N}\mathbb{E}[\|\bar{\mathbf{x}}_{k}-\mathbf{x}_{k}^{i}\|^{2}]\right)
\end{split}
\end{equation}

Substituting ~(\ref{helper-t3}),~(\ref{helper-t4}) into~(\ref{helper-t2}), we have 

\begin{equation}
\label{eq-t2}
\begin{split}
-\frac{\eta}{1-\beta}\mathbb{E}[\langle\nabla\mathcal{F}(\bar{\mathbf{x}}_{k}),\frac{1}{N}\sum_{i=1}^{N}\tilde{\mathbf{g}}^i_{k}\rangle]&\leq
\Big(\frac{(1-\beta)}{2 \beta L}-\frac{\eta}{2(1-\beta)}\Big)\mathbb{E}[\|\nabla \mathcal{F}(\bar{\mathbf{x}}_{k})\|^{2}]+
\frac{\eta^2 \beta L }{2(1-\beta)^{3}}\mathbb{E}[\|\frac{1}{N} \sum_{i=1}^{N} (\tilde{\mathbf{g}}_{k}^{i}- \mathbf{g}_k^i)\|^{2}]\\
&-\frac{\eta}{2(1-\beta)}\mathbb{E}[\|\frac{1}{N} \sum_{i=1}^{N} \nabla f_{i}(\mathbf{x}_{k}^{i})\|^{2}]+ \frac{\eta L^2}{2(1-\beta)} \frac{1}{N} \sum_{i=1}^{N}\mathbb{E}[\|\bar{\mathbf{x}}_{k}-\mathbf{x}_{k}^{i}\|^{2}]
\end{split}
\end{equation}

Now, finally substituting ~(\ref{eq-t2}),~(\ref{eq-t1}) into~(\ref{main1}), we have

\begin{equation}
\label{main2}
\begin{split}
&\mathbb{E}[\langle\nabla\mathcal{F}(\bar{\mathbf{z}}_{k}),\bar{\mathbf{z}}_{k+1}- \bar{\mathbf{z}}_{k}\rangle] \leq 
 \frac{(1-\beta)L}{2\beta}\mathbb{E}[\|\bar{\mathbf{z}}_{k}-\bar{\mathbf{x}}_{k}\|^2]+ 
 \frac{\beta L \eta^2}{2(1-\beta)^3}\mathbb{E}[\|\frac{1}{N}\sum_{i=1}^{N}(\tilde{\mathbf{g}}^i_{k})\|^2]+
 \bigg(\frac{(1-\beta)}{2\beta L}- \frac{\eta}{2(1-\beta)}\bigg)\\
 &\mathbb{E}[\|\nabla \mathcal{F}(\bar{\mathbf{x}}_k)\|^2 ] - \frac{\eta}{2(1-\beta)}\mathbb{E}[\|\frac{1}{N}\sum_{i=1}^{N}\nabla f_i(\mathbf{x}^i_k)\|^2]+\frac{\eta^2 \beta L}{2(1-\beta)^3}\mathbb{E}[\|\frac{1}{N}\sum_{i=1}^{N}(\tilde{\mathbf{g}}^i_{k}-\mathbf{g}^i_k)\|^2 ]+\frac{\eta L^2}{2(1-\beta)} \frac{1}{N}\sum_{i=1}^{N}\mathbb{E}[\|\bar{\mathbf{x}}_{k}-\mathbf{x}^i_k\|^2]
\end{split}
\end{equation}

Equation.~\ref{fact_1} states that:
\begin{equation}\label{eqlem_3}
    \mathbb{E}[\|\bar{\mathbf{z}}_{k+1}-\bar{\mathbf{z}}_k\|^2]=\frac{\eta^2}{(1-\beta)^2}\mathbb{E}[\|\frac{1}{N}\sum_{i=1}^N\tilde{\mathbf{g}}^i_k\|^2].
\end{equation}

Substituting (\ref{main2}), (\ref{eqlem_3}) in (\ref{main0}):

\begin{equation*}
\begin{split}
    &\mathbb{E}[\mathcal{F}(\bar{\mathbf{z}}_{k+1})] \leq \mathbb{E}[\mathcal{F}(\bar{\mathbf{z}}_{k})]+ \frac{(1-\beta)L}{2\beta}\mathbb{E}[\|\bar{\mathbf{z}}_{k}-\bar{\mathbf{x}}_{k}\|^2]+
     \frac{\beta L \eta^2}{2(1-\beta)^3}\mathbb{E}[\|\frac{1}{N}\sum_{i=1}^{N}(\tilde{\mathbf{g}}^i_{k})\|^2]+
     \bigg(\frac{(1-\beta)}{2\beta L}- \frac{\eta}{2(1-\beta)}\bigg)\\
 &\mathbb{E}[\|\nabla \mathcal{F}(\bar{\mathbf{x}}_k)\|^2 ] - \frac{\eta}{2(1-\beta)}\mathbb{E}[\|\frac{1}{N}\sum_{i=1}^{N}\nabla f_i(\mathbf{x}^i_k)\|^2]+\frac{\eta^2 \beta L}{2(1-\beta)^3}\mathbb{E}[\|\frac{1}{N}\sum_{i=1}^{N}(\tilde{\mathbf{g}}^i_{k}-\mathbf{g}^i_k)\|^2 ]+\\
 &\frac{\eta L^2}{2(1-\beta)} \frac{1}{N}\sum_{i=1}^{N}\mathbb{E}[\|\bar{\mathbf{x}}_{k}-\mathbf{x}^i_k\|^2] +\frac{\eta^2}{(1-\beta)^2}\mathbb{E}[\|\frac{1}{N}\sum_{i=1}^N\tilde{\mathbf{g}}^i_k\|^2].
    \end{split}
\end{equation*}

We find the bound for $\mathbb{E}[\|\nabla\mathcal{F}(\bar{\mathbf{x}}_{k})\|^2]$ by rearranging the terms and dividing by $C_1 = \frac{\eta}{2(1-\beta)}-\frac{(1-\beta)}{2\beta L}$:

\begin{equation*}
\begin{split}
    &\mathbb{E}[\|\nabla\mathcal{F}(\bar{\mathbf{x}}_{k})\|^2] \leq
    \frac{1}{C_1}\bigg(\mathbb{E}[\mathcal{F}(\bar{\mathbf{z}}_{k})]-\mathbb{E}[\mathcal{F}(\bar{\mathbf{z}}_{k+1})]\bigg)+
    C_2 \:\mathbb{E}[\|\frac{1}{N}\sum_{i=1}^{N}(\tilde{\mathbf{g}}^i_{k})\|^2]+
    C_3\:\mathbb{E}[\|\bar{\mathbf{z}}_{k}-\bar{\mathbf{x}}_{k}\|^2]\\
    &- C_6\: \mathbb{E}[\|\frac{1}{N}\sum_{i=1}^{N}\nabla f_i(\mathbf{x}^i_k)\|^2] + C_4\:\mathbb{E}[\|\frac{1}{N}\sum_{i=1}^{N}(\tilde{\mathbf{g}}^i_{k}-\mathbf{g}^i_k)\|^2 ] + \frac{C_5}{N}\: \sum_{i=1}^{N}\mathbb{E}[\|\bar{\mathbf{x}}_{k}-\mathbf{x}^i_k\|^2]
    \end{split}
\end{equation*}
Where $C_{2}=\left(\frac{\beta L \eta^{2}}{2(1-\beta)^{3}}+\frac{\eta^{2} L}{(1-\beta)^{2}}\right) / C_{1}$, $C_{3}=\frac{(1-\beta) L}{2 \beta} / C_{1}$,  $C_{4}=\frac{\eta^2 \beta L}{2(1-\beta)^{3}} / C_{1}$,  $C_{5}=\frac{\eta L^{2}}{2(1-\beta)} / C_{1}$,
$C_{6}=\frac{\eta}{2(1-\beta)} / C_{1}$.\\

Summing over $k \in \{0,1,\dots, K-1\}$: 

\begin{equation*}
    \begin{split}
        &\sum_{k=0}^{K-1} \mathbb{E}\left[\left\|\nabla \mathcal{F}\left(\bar{\mathbf{x}}_{k}\right)\right\|^{2}\right] \leq
        \frac{1}{C_{1}}\bigg(\mathbb{E}\left[\mathcal{F}\left(\bar{\mathbf{z}}_{0}\right)\right]-\mathbb{E}\left[\mathcal{F}\left(\bar{\mathbf{z}}_{k}\right)\right]\bigg)-C_{6} \sum_{k=0}^{K-1}
        \mathbb{E}\left[\left\|\frac{1}{N} \sum_{i=1}^{N} \nabla f_{i}\left(\mathbf{x}_{k}^{i}\right)\right\|^{2}\right]+ C_{2} \sum_{k=0}^{K-1} \mathbb{E}\left[\left\|\frac{1}{N} \sum_{i=1}^{N} \tilde{\mathbf{g}}_{k}^{i}\right\|^{2}\right]\\ & + C_{3} \sum_{k=0}^{K-1} \mathbb{E}\left[\left\|\bar{\mathbf{z}}_{k}-\bar{\mathbf{x}}_{k}\right\|^{2}\right]+C_{4} \sum_{k=0}^{K-1} \mathbb{E}\left[\left\|\frac{1}{N} \sum_{i=1}^{N}\left(\tilde{\mathbf{g}}_{k}^{i}-\mathbf{g}_{k}^{i}\right)\right\|^{2}\right] +C_{5} \sum_{k=0}^{K-1} \frac{1}{N} \sum_{l=1}^{N} \mathbb{E}\left[\left\|\bar{\mathbf{x}}_{k}-\mathbf{x}_{k}^{i}\right\|^{2}\right]
    \end{split}
\end{equation*}

Substituting Lemma~\ref{lemma1}, Lemma~\ref{lemma_2}, Lemma~\ref{lemma_3} and Equation.~\ref{fact_2} into the above equation, we have:

\begin{equation*}
    \begin{split}
        &\sum_{k=0}^{K-1} \mathbb{E}\left[\left\|\nabla \mathcal{F}\left(\bar{\mathbf{x}}_{k}\right)\right\|^{2}\right] \leq
        \frac{1}{C_{1}}\bigg(\mathbb{E}\left[\mathcal{F}\left(\bar{\mathbf{z}}_{0}\right)\right]-\mathbb{E}\left[\mathcal{F}\left(\bar{\mathbf{z}}_{k}\right)\right]\bigg)-
        \bigg(C_{6}-2C_2-2C_3\frac{\eta^2 \beta^2}{(1-\beta)^4}-C_5\frac{6\eta^2}{(1-\beta)^2(1-\sqrt{\rho})}\bigg)\\ 
        &\sum_{k=0}^{K-1}
        \mathbb{E}\left[\left\|\frac{1}{N} \sum_{i=1}^{N} \nabla f_{i}\left(\mathbf{x}_{k}^{i}\right)\right\|^{2}\right]+ 
        \bigg(C_{2}+C_3\frac{\eta^2\beta^2}{(1-\beta)^4}\bigg)\bigg(\frac{10\sigma^2}{N}+8\zeta^2\bigg)K
        + 4C_4\bigg(\frac{\sigma^2}{N}+\zeta^2\bigg)K\\
        &+ C_5 \frac{14\eta^2(\sigma^2+\zeta^2)}{(1-\beta)^2(1-\sqrt{\rho})^2}K
    \end{split}
\end{equation*}

Dividing both sides by $K$ and considering the fact that $\bar{\mathbf{z}}_0 = \bar{\mathbf{x}}_0$ and $ \bigg(C_{6}-2C_2-2C_3\frac{\eta^2 \beta^2}{(1-\beta)^4}-C_5\frac{6\eta^2}{(1-\beta)^2(1-\sqrt{\rho})}\bigg) \geq 0$:
\begin{equation*}
    \begin{split}
        \frac{1}{K}\sum_{k=0}^{K-1} \mathbb{E}\left[\left\|\nabla \mathcal{F}\left(\bar{\mathbf{x}}_{k}\right)\right\|^{2}\right]& \leq
        \frac{1}{C_{1}K}\bigg(\mathcal{F}\left(\bar{\mathbf{x}}_{0}\right)-\mathcal{F}^{\star}\bigg)+
         \bigg(C_{2}+C_3\frac{\eta^2\beta^2}{(1-\beta)^4}\bigg)\bigg(\frac{10\sigma^2}{N}+8\zeta^2\bigg)\\
        &+ 4C_4\bigg(\frac{\sigma^2}{N}+\zeta^2\bigg) + C_5 \frac{14\eta^2(\sigma^2+\zeta^2)}{(1-\beta)^2(1-\sqrt{\rho})^2}\\
        & \leq
        \frac{1}{C_{1}K}\bigg(\mathcal{F}\left(\bar{\mathbf{x}}_{0}\right)-\mathcal{F}^{\star}\bigg)+
         \bigg(10C_{2}+10C_3\frac{\eta^2\beta^2}{(1-\beta)^4}+ 4C_4\bigg)\bigg(\frac{\sigma^2}{N}+\zeta^2\bigg)\\
        & + C_5 \frac{14\eta^2(\sigma^2+\zeta^2)}{(1-\beta)^2(1-\sqrt{\rho})^2}
    \end{split}
\end{equation*}

Therefore we arrive that the bound given by the theorem.~\ref{theorem_1}:
\begin{equation}
\label{eq_them1}
    \begin{split}
        \frac{1}{K}\sum_{k=0}^{K-1}  \mathbb{E}[||\nabla  \mathcal{F}(\Bar{x}_k)||^2] &\leq \frac{1}{C_1K}(\mathcal{F}(\Bar{x}^0)-\mathcal{F}^*) +
         \bigg(10C_{2}+10C_3\frac{\eta^2\beta^2}{(1-\beta)^4} + 4C_4\bigg)(\frac{\sigma^2}{N}+\zeta^2) +\\
         & C_5 \frac{14\eta^2(\sigma^2+\zeta^2)}{(1-\beta)^2(1-\sqrt{\rho})^2}\\
    \end{split}
\end{equation}

\subsubsection{Discussion on the Step Size}
\label{step-size}
In the proof of Theorem~\ref{theorem_1}, we assumed the following
$ C_{6}-2C_2-2C_3\frac{\eta^2 \beta^2}{(1-\beta)^4}-C_5\frac{6\eta^2}{(1-\beta)^2(1-\sqrt{\rho})} \geq 0$.\\
The above equation is true under the following conditions:
\begin{equation*}
\begin{split}
    (i)  & \hspace{2mm} C_1 > 0\\
    (ii) &\hspace{2mm} 1-\frac{4L}{(1-\beta)^2} \eta - \frac{6L^2}{(1-\beta)^2 (1-\sqrt{\rho})} \eta^2 \geq 0
\end{split}
\end{equation*}
Solving the first inequality gives us $\frac{(1-\beta)^2}{\beta L} < \eta$.

Now, solving the second inequality, combining the fact that $\eta>0$, we have then the specific form of $\eta^*$
\[\eta^*=\frac{\sqrt{4(1-\sqrt{\rho})^2+6(1-\sqrt{\rho})(1-\beta)^2}-2(1-\sqrt{\rho})}{6L}.\]
Therefore, the step size $\eta$ is defined as 
\[\frac{(1-\beta)^2}{\beta L} < \eta \leq \frac{\sqrt{4(1-\sqrt{\rho})^2+6(1-\sqrt{\rho})(1-\beta)^2}-2(1-\sqrt{\rho})}{6L}\]

\subsubsection{Proof for Corollary 1}
\label{coro_1_proof}
We assume that the step size $\eta$ is $\mathcal{O}(\frac{\sqrt{N}}{\sqrt{K}})$ and $\zeta^2$ is $\mathcal{O}(\frac{1}{\sqrt{K}})$.  Given this assumption, we have the following
\[C_1=\mathcal{O}(\frac{\sqrt{N}}{\sqrt{K}}), C_2=\mathcal{O}(\frac{\sqrt{N}}{\sqrt{K}}),  C_3=\mathcal{O}(\frac{\sqrt{K}}{\sqrt{N}}), C_4=\mathcal{O}(\frac{\sqrt{N}}{\sqrt{K}}), C_5=\mathcal{O}(1),\] 
Now we proceed to find the order of each term in Equation.~\ref{eq_them1}. To do that we first point out that
\[\frac{\mathcal{F}(\bar{\mathbf{x}}_0)-\mathcal{F}^*}{C_1K}=\mathcal{O}\Big(\frac{1}{\sqrt{NK}}\Big).\]
For the remaining terms we have,
\[\bigg(10C_{2}+10C_3\frac{\eta^2\beta^2}{(1-\beta)^4} + 4C_4\bigg)\frac{\sigma^2}{N} =\mathcal{O}\Big(\frac{1}{\sqrt{NK}}\Big), \hspace{3mm} \bigg(10C_{2}+10C_3\frac{\eta^2\beta^2}{(1-\beta)^4} + 4C_4\bigg)\zeta^2=\mathcal{O}\Big(\frac{\sqrt{N}}{K}\Big)\]

\[C_5 \frac{14\eta^2\sigma^2}{(1-\beta)^2(1-\sqrt{\rho})^2} = \mathcal{O}\Big(\frac{N}{K}\Big) , \hspace{3mm} C_5 \frac{14\eta^2\zeta^2}{(1-\beta)^2(1-\sqrt{\rho})^2} = \mathcal{O}\Big(\frac{N}{K^{1.5}}\Big) \]

Therefore, by omitting the constant $N$ in this context, there exists a constant $C>0$ such that the overall convergence rate is as follows:
\begin{equation}
    \frac{1}{K}\sum_{k=0}^{K-1} \mathbb{E}\left[\left\|\nabla \mathcal{F}\left(\bar{\mathbf{x}}_{k}\right)\right\|^{2}\right] \leq C\Bigg(\frac{1}{\sqrt{NK}}+\frac{1}{K}+\frac{1}{K^{1.5}}\Bigg), 
\end{equation}
which suggests when $N$ is fixed and $K$ is sufficiently large, \textit{NGC} enables the convergence rate of $\mathcal{O}(\frac{1}{\sqrt{NK}})$.

\subsection{{Decentralized Learning Setup}}
\label{apx:dl}
{The traditional decentralized learning algorithm (d-psgd) is described as Algorithm.~\ref{alg:dl}. For the decentralized setup, we use an undirected ring and undirected torus graph topologies with a uniform mixing matrix. The undirected ring topology for any graph size has 3 peers per agent including itself and each edge has a weight of $\frac{1}{3}$. The undirected torus topology with 10 agents has 4 peers per agent including itself and each edge has a weight of $\frac{1}{4}$.
The undirected torus topology 20 agents have 5 peers per agent including itself and each edge has a weight of $\frac{1}{5}$.}

\begin{algorithm}[ht]
\textbf{Input:} Each agent $i \in [1,N]$ initializes model weights $x^i_{(0)}$, step size $\eta$, averaging rate $\gamma$, mixing matrix $W=[w_{ij}]_{i,j \in [1,N]}$, and $I_{ij}$ are elements of $N\times N$ identity matrix.\\

Each agent simultaneously implements the 
T\text{\scriptsize RAIN}( ) procedure\\
1.  \textbf{procedure} T\text{\scriptsize RAIN}( ) \\
2.  \hspace{4mm}\textbf{for} k=$0,1,\hdots,K-1$ \textbf{do}\\
3.  \hspace*{8mm}$d^i_{k} \sim D^i$\hfill\textcolor{gray!80}{// sample data from training dataset.}\\
4.  \hspace*{8mm}$g^{i}_{k}=\nabla_x f_i(d^i_{k}; x_i^{k}) $ \hfill \textcolor{gray!80}{// compute the local gradients}\\
5. \hspace*{8mm}$v^i_{k}= \beta v^i_{(k-1)} - \eta g^i_{k}$\hfill \textcolor{gray!80}{// momentum step}\\
6.  \hspace*{8mm}$\widetilde{x}^i_{k}=x^i_{k}+v^i_{k}$\hfill\textcolor{gray!80}{// update the model}\\
7.  \hspace*{8mm}S\text{\scriptsize END}R\text{\scriptsize ECEIVE}($\widetilde{x}^i_{k}$)\hfill\textcolor{gray!80}{// share model parameters with neighbors $N(i)$.}\\
8.  \hspace*{8mm}$x^i_{(k+1)}=\widetilde{x}^i_{k}+\gamma\sum_{j\in \mathcal{N}(i)}(w_{ij}-I_{ij})*\widetilde{x}^j_{k}$\hfill \textcolor{gray!80}{// gossip averaging step}\\
9.  \hspace{4mm}\textbf{end}\\
10.  \textbf{return}
\caption{{Decentralized Peer-to-Peer Training (\textit{D-PSGD} with momentum)}}
\label{alg:dl}
\end{algorithm}

\subsection{{CompNGC Algorithm}}
The section presents the pseudocode for Compressed \textit{NGC} in Algorithm~\ref{apx_alg:compNGC}
\begin{algorithm}[t]
\textbf{Input:} Each agent $i \in [1,N]$ initializes model weights $x^i_{(0)}$, step size $\eta$, averaging rate $\gamma$, dimension of the gradient $d$, mixing matrix $W=[w_{ij}]_{i,j \in [1,N]}$, \textit{NGC} mixing weight $\alpha$, and $I_{ij}$ are elements of $N\times N$ identity matrix.\\

Each agent simultaneously implements the 
T\text{\scriptsize RAIN}( ) procedure\\
1.  \textbf{procedure} T\text{\scriptsize RAIN}( ) \\
2.  \hspace{4mm}\textbf{for} k=$0,1,\hdots,K-1$ \textbf{do}\\
3.  \hspace*{8mm}$d^i_{k} \sim D^i$\hfill\textcolor{gray!80}{// sample data from training dataset}\\
4.  \hspace*{8mm}$g^{ii}_{k}=\nabla_x f_i(d^i_{k}; x^i_{k}) $ \hfill \textcolor{gray!80}{// compute the local self-gradients}\\
5.  \hspace*{8mm}$p^{ii}_{k}=g^{ii}_{k}+e^{ii}_{k}$\hfill \textcolor{gray!80}{// error compensation for self-gradients}\\ 
6.  \hspace*{8mm}$\delta^{ii}_{k}=(||p^{ii}_{k}||_1/d)sgn(p^{ii})$\hfill\textcolor{gray!80}{// compress the compensated self-gradients}\\
7.  \hspace*{8mm}$e^{ii}_{k+1}=p^{ii}_{k}-\delta^{ii}_{k}$\hfill\textcolor{gray!80}{// update the error variable}\\
8.  \hspace*{8mm}S\text{\scriptsize END}R\text{\scriptsize ECEIVE}($x^i_{k}$)\hfill\textcolor{gray!80}{// share model parameters with neighbors $N(i)$}\\
9.  \hspace*{8mm}\textbf{for} each neighbor $j \in \{N(i)-i\}$ \textbf{do}\\
10. \hspace*{12mm}$g^{ji}_{k}=\nabla_x f_i(d^i_{k}; x^j_{k})$\hfill \textcolor{gray!80}{// compute neighbors' cross-gradients}\\
11. \hspace*{12mm}$p^{ji}_{k}=g^{ji}_{k}+e^{ji}_{k}$\hfill\textcolor{gray!80}{// error compensation for cross-gradients}\\
12. \hspace*{12mm}$\delta^{ji}_{k}=(||p^{ji}_{k}||_1/d)sgn(p^{ji})$\hfill\textcolor{gray!80}{// compress the compensated cross-gradients}\\
13. \hspace*{12mm}$e^{ji}_{k+1}=p^{ji}_{k}-\delta^{ji}_{k}$\hfill\textcolor{gray!80}{// update the error variable}\\
14. \hspace*{12mm}\textbf{if} $\alpha \neq 0$ \textbf{do}\\
15. \hspace*{16mm}S\text{\scriptsize END}R\text{\scriptsize ECEIVE}($\delta^{ji}_{k}$)\hfill\textcolor{gray!80}{// share compressed cross-gradients between $i$ and $j$}\\
16. \hspace*{12mm}\textbf{end}\\
17. \hspace*{8mm}\textbf{end}\\
18. \hspace*{8mm}$\widetilde{g}^i_{k}=(1-\alpha)*\sum_{j\in \mathcal{N}(i)}w_{ji}*\delta^{ji}_{k}+ \alpha*\sum_{j\in \mathcal{N}(i)}w_{ij}*\delta^{ij}_{k}$\hfill\textcolor{gray!80}{// modify local gradients}\\
19.  \hspace*{8mm}$v^i_{k}= \beta v^i_{(k-1)} - \eta \widetilde{g}^i_{k}$\hfill \textcolor{gray!80}{// momentum step}\\
20. \hspace*{8mm}$\widetilde{x}^i_{k}=x^i_{k}+v^i_{k}$\hfill \textcolor{gray!80}{// update the model} \\
21. \hspace*{8mm}$x^i_{(k+1)}=\widetilde{x}^i_{k}+\gamma\sum_{j\in \mathcal{N}(i)}(w_{ij}-I_{ij})*x^j_{k}$\hfill\textcolor{gray!80}{// gossip averaging step}\\
22. \hspace*{4mm}\textbf{end}\\
23. \textbf{return}
\caption{Compressed Neighborhood Gradient Clustering (\textit{CompNGC})}
\label{apx_alg:compNGC}
\end{algorithm}

\subsection{{Datasets}}
\label{apx:datasets}
{In this section, we give a brief description of the datasets used in our experiments. We use a diverse set of datasets each originating from a different distribution of images to show the generalizability of the proposed techniques.}

\textbf{CIFAR-10:} 
{CIFAR-10 \cite{cifar} is an image classification dataset with 10 classes. The image samples are colored (3 input channels) and have a resolution of $32 \times 32$. There are $50,000$ training samples with $5000$ samples per class and $10,000$ test samples with $1000$ samples per class.}

\textbf{CIFAR-100:} 
{CIFAR-100 \cite{cifar} is an image classification dataset with 100 classes. The image samples are colored (3 input channels) and have a resolution of $32 \times 32$. There are $50,000$ training samples with $500$ samples per class and $10,000$ test samples with $100$ samples per class. CIFAR-100 classification is a harder task compared to CIFAR-10 as it has 100 classes with very less samples per class to learn from.}

\textbf{Fashion MNIST:}
{Fashion MNIST \cite{fmnist} is an image classification dataset with 10 classes. The image samples are in greyscale (1 input channel) and have a resolution of $28 \times 28$. There are $60,000$ training samples with $6000$ samples per class and $10,000$ test samples with $1000$ samples per class.}

\textbf{Imagenette:}
{Imagenette \cite{imagenette} is a 10-class subset of the ImageNet dataset. The image samples are in colored (3 input channels) and have a resolution of $224 \times 224$. There are $9469$ training samples with roughly $950$ samples per class and $3925$ test samples. We conduct our experiments on two different resolutions of the Imagenette dataset -- a) a resized low resolution of $32 \times 32$ and, b) a full resolution of $224 \times 224$. The Imagenette experimental results reported in Table.~\ref{tab:datasets} use the low-resolution images whereas experimental results in Table.~\ref{tab:nlp-fr_imagenette} use the full resolution images.}

\textbf{AGNews:} { We use AGNews \cite{agnews} dataset for Natural Language Processing (NLP) task. This is a text classification dataset where the given text news is classified into 4 classes, namely "World", "Sport", "Business" and "Sci/Tech". The dataset has a total of 120000 and 7600 samples for training and testing respectively, which are equally distributed across each class. } 

\subsection{Network Architecture}
\label{apx:arch}
{We replace ReLU+BatchNorm layers of all the model architectures with EvoNorm-S0 \cite{evonorm} as it was shown to be better suited for decentralized learning over non-IID distributions \cite{qgm}.}

\textbf{5 layer CNN:} The 5-layer CNN consists of 4 convolutional with EvoNorm-S0 \cite{evonorm} as activation-normalization layer, 3 max-pooling layers, and one linear layer. In particular, it has 2 convolutional layers with 32 filters, a max pooling layer, then 2 more convolutional layers with 64 filters each followed by another max pooling layer and a dense layer with 512 units. It has a total of $76K$ trainable parameters.

\textbf{VGG-11:} We modify the standard VGG-11 \cite{vgg} architecture by reducing the number of filters in each convolutional layer by $4\times$ and using only one dense layer with 128 units. 
Each convolutional layer is followed by EvoNorm-S0 as the activation-normalization layer and it has $0.58M$ trainable parameters.

\textbf{ResNet-20:} For ResNet-20 \cite{resnet}, we use the standard architecture with $0.27M$ trainable parameters except that BatchNorm+ReLU layers are replaced by EvoNorm-S0.

\textbf{LeNet-5:} {For LeNet-5 \cite{lenet}, we use the standard architecture with $61,706$ trainable parameters.}

\textbf{MobileNet-V2:} {We use the the standard MobileNet-V2 \cite{mobilnetv2} architecture used for CIFAR dataset with $2.3M$ parameters except that BatchNorm+ReLU layers are replaced by EvoNorm-S0.}

\textbf{ResNet-18:} {For ResNet-18 \cite{resnet}, we use the standard architecture with $11M$ trainable parameters except that BatchNorm+ReLU layers are replaced by EvoNorm-S0.}

\textbf{BERT\textsubscript{mini}:} {For BERT\textsubscript{mini} \cite{bert} we use the standard model from the paper. We restrict the sequence length of the model to 128. The model used in the work hence has $11.07M$ parameters. }

\textbf{DistilBERT\textsubscript{base}:} {For DistilBERT\textsubscript{base} \cite{distilbert} we use the standard model from the paper. We restrict the sequence length of the model to 128. The model used in the work hence has $66.67M$ parameters. }

\begin{figure}[ht]
\centering
\begin{subfigure}{0.32\textwidth}
    \vspace{-4mm}
    \includegraphics[width=\textwidth]{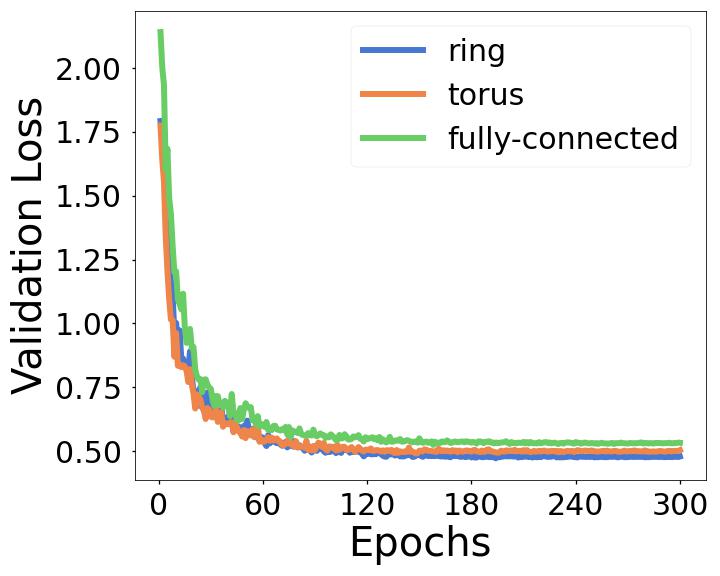}
    \caption{\textit{NGC} method on IID data.}
    \label{fig:iid_curve_10agents}
\end{subfigure}
\hfill
\begin{subfigure}{0.32\textwidth}
 \vspace{-4mm}
    \includegraphics[width=\textwidth]{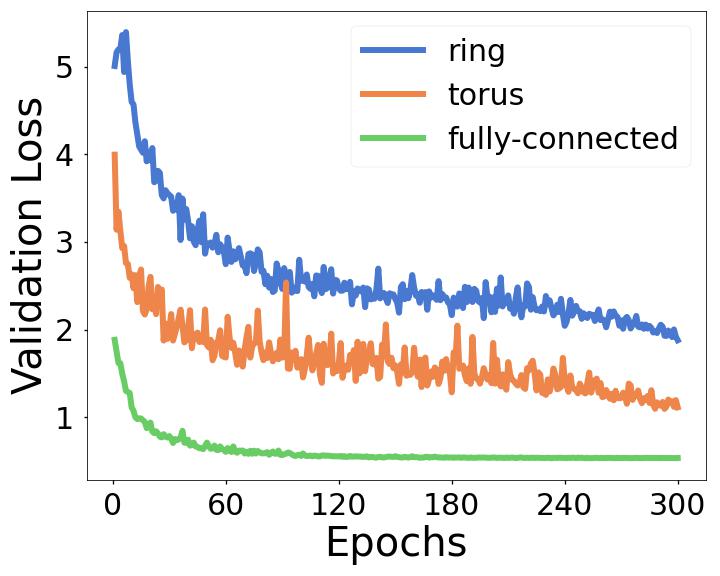}
    \caption{\textit{NGC} method on Non-IID data.}
    \label{fig:noniid_curve_10agents}
\end{subfigure}
\hfill
\begin{subfigure}{0.32\textwidth}
    \includegraphics[width=\textwidth]{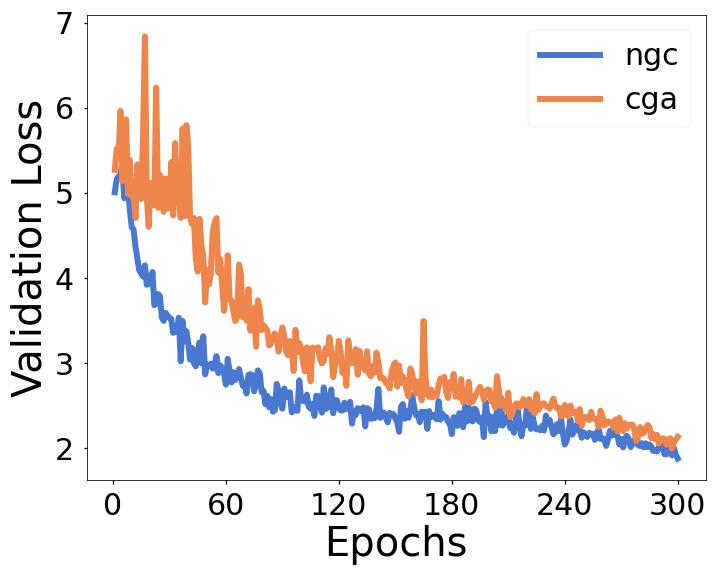}
    \caption{Different methods on Non-IID for ring topology.}
    \label{fig:algo_curve_10agents}
\end{subfigure}
        
\caption{Average Validation loss during training of 10 agents on CIFAR-10 dataset with a 5 layer CNN network.}
\label{fig:figures_10agents}
\end{figure}

\subsection{Analysis for 10 Agents}
\label{apx:10agent}
\begin{figure*}[ht]
     \centering
 \begin{subfigure}[t]{0.49\linewidth}
     \centering
     \includegraphics[width=\linewidth]{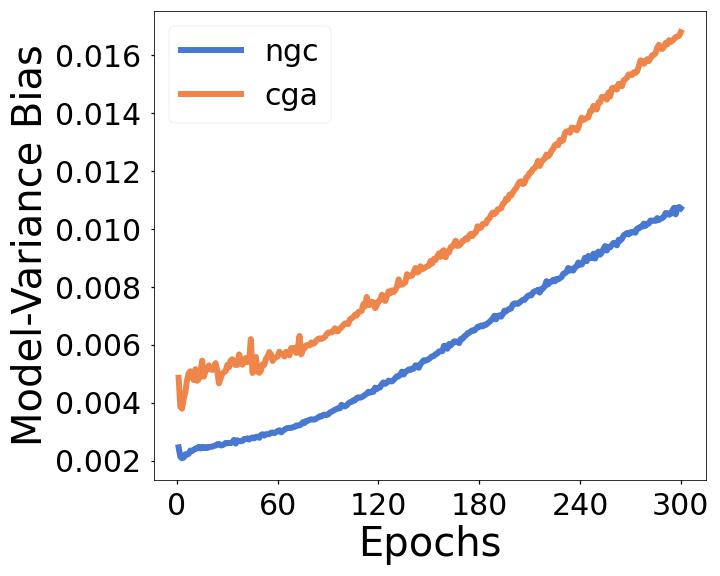}
     \caption{Average L1 norm of $\epsilon$.}
     \label{fig:epsilon_10agents}
 \end{subfigure}
 \begin{subfigure}[t]{0.49\linewidth}
     \centering
     \includegraphics[width=\linewidth]{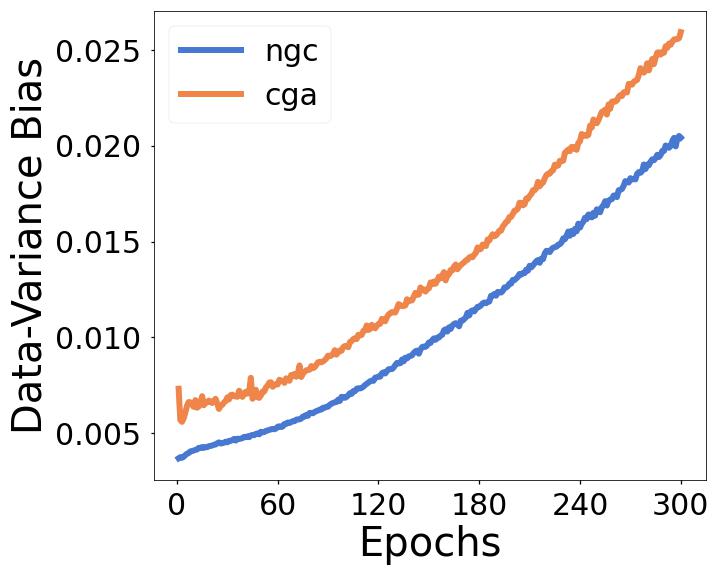}
     \caption{Average L1 norm of $\omega$.}
     \label{fig:omega_10agents}
 \end{subfigure}
    \caption{ Average L1 norm of model variance bias and data variance bias for 10 agents trained on CIFAR-10 dataset with 5 layer CNN architecture over an undirected ring topology.}
    \label{fig:norms_10agents}
\end{figure*}

We show the convergence characteristics of the proposed \textit{NGC} algorithm over IID and Non-IID data sampled from CIFAR-10 dataset in Figure.~\ref{fig:iid_curve_10agents}, and \ref{fig:noniid_curve_10agents} respectively. For Non-IID distribution, we observe that there is a slight difference in convergence rate (as expected) with slower rate for sparser topology (undirected ring graph) compared to its denser counterpart (fully connected graph). Figure.~\ref{fig:algo_curve_10agents} shows the comparison of the convergence characteristics of the \textit{NGC} technique with the current state-of-the-art CGA algorithm. We observe that \textit{NGC} has lower validation loss than CGA for same decentralized setup indicating its superior performance over CGA. We also plot the model variance and data variance bias terms for both \textit{NGC} and CGA techniques as shown in Figure.~\ref{fig:epsilon_10agents}, and \ref{fig:omega_10agents} respectively. 
We observe that both model variance and data variance bias for \textit{NGC} are significantly lower than CGA. 

\subsection{Resource Comparison} 
\label{apx:hardware}
The communication cost, memory overhead and compute overhead for various decentralized algorithms are shown in Table.~\ref{apx_tab:hardware}. The D-PSGD algorithm requires each agent to communicate model parameters of size $m_s$ with all the $N_i$ neighbors for the gossip averaging step and hence has a communication cost of $\mathcal{O}(m_sN_i)$. In the case of \textit{NGC} and CGA, there is an additional communication round for sharing data-variant cross gradients apart from sharing model parameters for the gossip averaging step. So, both these techniques incur a communication cost of $\mathcal{O}(2m_sN_i)$ and therefore an overhead of $\mathcal{O}(m_sN_i)$ compared to D-PSGD. \textit{CompNGC} compresses the additional round of communication involved with \textit{NGC} from $b$ bits to $1$ bit. This reduces the communication overhead from  $\mathcal{O}(m_sN_i)$ to  $\mathcal{O}(\frac{m_sN_i}{b})$. 

CGA algorithm stores all the received data-variant cross-gradients in the form of a matrix for quadratic projection step. Hence, CGA has a memory overhead of $\mathcal{O}(m_sN_i)$ compared to D-PSGD.
\textit{NGC} does not require any additional memory as it averages the received data-variant cross-gradients into self-gradient buffer.
The compressed version of \textit{NGC} requires an additional memory of $\mathcal{O}(m_sN_i)$ to store the error variables $e_{ji}$ (refer Algorith.~\ref{apx_alg:compNGC}). 
CompCGA also needs to store error variables along with the projection matrix of compressed gradients. Therefore, CompCGA has a memory overhead of $\mathcal{O}(m_sN_i+\frac{m_sN_i}{b})$. Note that memory overhead depends on the type of graph topology and model architecture but not on the size of the graph. The memory overhead for different model architectures trained on undirected ring topology is shown in Table.~\ref{tab:memory}

{The computation of the cross-gradients (in both CGA and NGC algorithms) requires $N_i$ forward and backward passes through the deep learning model at each agent. This is reflected as $\mathcal{O}(3N_iFP)$ in the compute overhead in Table.~\ref{apx_tab:hardware}. }
We assume that the compute effort required for the backward pass is twice that of the forward pass. 
CGA algorithm involves quadratic programming projection step \cite{qp} to update the local gradients. Quadratic programming solver (\textcolor{blue!80}{\href{https://pypi.org/project/quadprog/}{quadprog}}) uses  Goldfarb/Idnani dual algorithm. CGA uses quadratic programming to solve the following (Equation~\ref{eq:qp} -see Equation 5a in \cite{cga}) optimization problem in an iterative manner:
\begin{equation}
\label{eq:qp}
\begin{split}
&\text{min}_u \hspace{4mm}\frac{1}{2}u^T G G^T u + g^T G^T u\\
&\text{s.t.} \hspace{8mm} u\geq 0\\
\end{split}
\end{equation}
where, G is the matrix containing cross-gradients, g is the self-gradient and the optimal gradient direction $g^*$ in terms of the optimal solution of the above equation $u^*$ is $g^* = G^Tu^* + g$. 
The above optimization takes multiple iterations which results in compute and time complexity to be of polynomial(degree$\geq 2$) order. In contrast, NGC involves simple averaging step that requires $O(m_sN_i)$ addition operations. 

\subsection{{Communication Cost:}} 
\label{apx:comm}
{In this section we present the communication cost per agent in terms of Gigabytes of data transferred during the entire training process (refer Tables.~\ref{tab:cf10_comm}, \ref{tab:datasets_comm}, \ref{tab:nlp_comm}, \ref{tab:imagenette_comm}). The D-PSGD and \textit{NGC} with $\alpha=0$ have the lowest communication cost ($1\times$). We emphasize that \textit{NGC} with $\alpha=0$ outperforms D-PSGD in decentralized learning over label-wise non-IID data for same communication cost. \textit{NGC} and CGA have $2 \times$ communication overhead compared to D-PSGD where as \textit{CompNGC} and CompCGA have $1.03 \times$ communication overhead compared to D-PSGD. The compressed version of \textit{NGC} and CGA compresses the second round of cross-gradient communication to 1 bit. We assume the full-precision cross-gradients to be of 32-bit precision and hence the \textit{CompNGC} reduces the communication cost by $32 \times$ compared to \textit{NGC}.}

\begin{table}[t]
\caption{Memory overheads for various methods trained on different model architectures with CIFAR-10 dataset over undirected ring topology with 2 neighbors per agent.}
\label{tab:memory}
\begin{center}
\begin{tabular}{|c|c|c|c|c|}
\hline
Architecture & CGA  & \textit{NGC} & CompCGA & \textit{CompNGC}\\
& (MB)&(MB)&(MB)&(MB)\\
 \hline
5 layer CNN  &0.58 &0 &0.58 &0.60\\
VGG-11 &4.42 &0 & 4.42&4.56\\
ResNet-20 & 2.28& 0&2.28 &2.15\\
\hline
\end{tabular}
\end{center}
\end{table}

\begin{table}[t]
\caption{{Communication costs per agent in GBs  for experiments in Table~\ref{tab:cf10}}}
\label{tab:cf10_comm}
\begin{center}
\begin{tabular}{l c c c c c}
\hline
Method& Agents& 5layer CNN& 5layer CNN&VGG-11 & ResNet-20\\
& & Ring & Torus & Ring & Ring \\
 \hline
 \hline
     D-PSGD &  5 & 17.75 &-& 270.64 &127.19\\
       and  &  10 & 8.92 & 13.38 & 135.86 & 63.84\\
     \textit{NGC} $\alpha=0$ & 20 &  4.50 & 68.48 & 32.18\\
     
  \hline
     CGA &  5 & 35.48& -& 541.05&254.27\\
     and &  10 & 17.81 & 26.72 & 271.50 & 127.59\\
     \textit{NGC} &   20 &8.98& 17.95 & 136.72 & 64.25 \\
     
\hline
   CompCGA&   5 &18.31& -& 279.09 &131.16 \\
    and  &  10 & 9.20&13.79 & 140.10 & 65.84\\
    \textit{CompNGC}  &   20 & 4.64& 9.28 & 70.61 & 33.18\\
 \hline
\end{tabular}
\end{center}
\end{table}

\begin{table}[t]
\caption{{Communication costs per agent in GBs  for experiments in Table~\ref{tab:datasets}}}
\label{tab:datasets_comm}
\begin{center}
\begin{tabular}{l c c c c }
\hline
Method& Agents& Fashion MNIST & CIFAR-100 & Imagenette \\
& & (LeNet-5) & (ResNet-20) & (MobileNet-V2)  \\
 \hline
 \hline
     D-PSGD and&5 &17.25 & 103.74& 103.12\\
    \textit{NGC} $\alpha=0$ &10  & 8.61& 51.89& 51.60\\
     
  \hline
    CGA and&5 &34.50 & 207.47 & 206.23\\
    \textit{NGC} (ours)&10 & 17.23&103.79 & 103.19\\
  \hline
    CompCGA and &5& 17.79&106.98 & 106.34\\
    \textit{CompNGC} (ours)&10 & 8.88& 53.52 &53.21 \\
 \hline
\end{tabular}
\end{center}
\end{table}

\begin{table}[t]
\caption{{Communication costs per agent in GBs for experiments in Table~\ref{tab:nlp-fr_imagenette} (right)}}
\label{tab:imagenette_comm}
\begin{center}
\begin{tabular}{l  c c }
\hline
Method&    Ring topology & chain topology \\
 \hline
 \hline
    D-PSGD and & 501.98 &401.59 \\
    \textit{NGC} $\alpha=0$ & &  \\
      \hline
    CGA and & 1003.96 & 803.17 \\
    \textit{NGC} & & \\
      \hline
    CompCGA and & 517.67 & 414.14\\
    \textit{CompNGC} & & \\
     \hline
\end{tabular}
\end{center}
\end{table}

\begin{table}[ht]
\caption{{Communication costs per agent in GBs for experiments in Table~\ref{tab:nlp-fr_imagenette} (left) }}
\label{tab:nlp_comm}
\begin{center}
\begin{tabular}{l  c c c }
\hline
Method&  \multicolumn{2}{c}{BERT\textsubscript{mini} }&  DistilBERT\textsubscript{base}\\
&   Agents = 4 &Agents = 8&   Agents = 4  \\
 \hline
 \hline
    D-PSGD and & 234.30&118.20 & 1410.39\\
    \textit{NGC} $\alpha=0$ & & & \\
      \hline
    CGA and &486.59 & 236.40 & 2820.77\\
    \textit{NGC} & & & \\
      \hline
    CompCGA and & 241.6& 121.89 & 1454.46\\
    \textit{CompNGC} & & & \\
     \hline
\end{tabular}
\end{center}
\end{table}

\subsection{Hyper-parameters}
\label{apx:hyperparameters}

{All the experiments were run for three randomly chosen seeds.}
{We decay the step size by 10x after 50\% and 75\% of the training, unless mentioned otherwise.}

\begin{table}[t]
\caption{Hyper-parameters used for CIFAR-10 with non-IID data distribution using 5-layer CNN model architecture presented in Table~\ref{tab:cf10}}
\label{tab:5cnn-hp}
\begin{center}
\begin{tabular}{c c c c}
\hline
& Agents& Ring&Torus\\
\cline{3-4}
Method & (n)& ($\alpha$, $\beta$, $\eta$, $\gamma$) & ($\alpha$, $\beta$, $\eta$, $\gamma$)\\
 \hline
     &  5 &($-, 0.0, 0.1, 1.0$) & $-$ \\
     D-PSGD &  10 &($-,0.0, 0.1, 1.0$) &($-,0.0, 0.1, 1.0$) \\
     &  20 & ($-,0.0, 0.1, 1.0$)& ($-,0.0, 0.1, 1.0$) \\
  \hline
  &  5 & ($0.0,0.0, 0.1, 1.0$)&$-$ \\
  \textit{NGC} (ours) &  10 &($0.0,0.0, 0.1, 1.0$) & ($0.0,0.0, 0.1, 1.0$) \\
   ($\alpha=0$)  &  20 &($0.0,0.0, 0.1, 1.0$) & ($0.0,0.0, 0.1, 1.0$) \\
\hline
     &  5 & ($-, 0.9, 0.01, 0.1$)& $-$\\
     CGA &  10 & ($-, 0.9, 0.01, 0.5$)&  ($-, 0.9, 0.01, 0.1$)\\
     &   20 &($-, 0.9, 0.01, 0.5$) &  ($-, 0.9, 0.01, 0.1$) \\
     \hline
 &  5 & ($1.0, 0.9, 0.01, 0.1$)&$-$ \\
    \textit{NGC} (ours) &  10 & ($1.0, 0.9, 0.01, 0.5$)& ($1.0, 0.9, 0.01, 0.1$) \\
     &  20 & ($1.0, 0.9, 0.01, 0.5$)& ($1.0, 0.9, 0.01, 0.1$) \\
\hline
   &   5 &($-, 0.9, 0.01, 0.1$) & $-$\\
    CompCGA &  10 &($-, 0.9, 0.01, 0.5$) & ($-, 0.9, 0.01, 0.1$) \\
     &   20 & ($-, 0.9, 0.01, 0.5$)&  ($-, 0.9, 0.01, 0.1$)\\
 \hline
  &  5 &($1.0, 0.9, 0.01, 0.1$) & $-$\\
   \textit{CompNGC} (ours) & 10 &($1.0, 0.9, 0.01, 0.5$) &($1.0, 0.9, 0.01, 0.1$) \\
     &  20 &($1.0, 0.9, 0.01, 0.5$) & ($1.0, 0.9, 0.01, 0.1$) \\
 \hline
\end{tabular}
\end{center}
\end{table}

\begin{table}[t]
\caption{Hyper-parameters used for CIFAR-10 with non-IID data distribution using ResNet and VGG-11 model architecture  presented in Table~\ref{tab:cf10}}
\label{tab:resnet-hp}
\begin{center}
\begin{tabular}{c c c c}
\hline
&Agents & VGG-11&ResNet\\
\cline{3-4}
Method & (n) & ($\alpha$, $\beta$, $\eta$, $\gamma$) & ($\alpha$, $\beta$, $\eta$, $\gamma$)\\
 \hline
     &  5 & ($-,0.0,0.01,1.0$)&($-,0.0, 0.1, 1.0$) \\
     D-PSGD &  10 & ($-,0.0,0.01,1.0$)&($-,0.0, 0.1, 1.0$) \\
     &  20 &($-,0.0,0.01,1.0$) & ($-,0.0, 0.1, 1.0$) \\
  \hline
  &  5 & ($0.0,0.0,0.01,1.0$)&($0.0,0.0, 0.1, 1.0$) \\
    \textit{NGC} (ours) &  10 & ($0.0,0.0,0.01,1.0$)& ($0.0,0.0, 0.1, 1.0$) \\
   ($\alpha=0$)  &  20 &($0.0,0.0,0.01,1.0$) & ($0.0,0.0, 0.1, 1.0$) \\
\hline
     &  5 & ($-,0.9,0.1,0.5$)&($-,0.9, 0.1, 1.0$) \\
     CGA &  10 & ($-,0.9,0.1,0.5$)&($-,0.9, 0.1, 1.0$)  \\
     &   20 & ($-,0.9,0.1,0.5$)& ($-,0.9, 0.1, 1.0$)  \\
     \hline
 &  5 & ($1.0,0.9,0.1,0.5$)& ($1.0,0.9, 0.1, 1.0$) \\
    \textit{NGC} (ours) &  10 & ($1.0,0.9,0.1,0.5$)& ($1.0,0.9, 0.1, 1.0$)  \\
     &  20 & ($1.0,0.9,0.1,0.5$)& ($1.0,0.9, 0.1, 1.0$)  \\
\hline
   &   5 & ($-,0.9,0.01,0.1$)&($-,0.9, 0.01, 0.1$) \\
    CompCGA &  10 & ($-,0.9,0.01,0.1$)& ($-,0.9, 0.01, 0.1$)\\
     &   20 & ($-,0.9,0.01,0.1$)& ($-,0.9, 0.01, 0.1$)\\
 \hline
  &  5 & ($1.0,0.9,0.01,0.1$)& ($1.0,0.9, 0.01, 0.1$) \\
   \textit{CompNGC} (ours) & 10 & ($1.0,0.9,0.01,0.1$)& ($1.0,0.9, 0.01, 0.1$)\\
     &  20 & ($1.0,0.9,0.01,0.1$)& ($1.0,0.9, 0.01, 0.1$) \\
 \hline
 \end{tabular}
\end{center}
\end{table}

\textbf{Hyper-parameters for CIFAR-10 on 5 layer CNN:} All the experiments that involve 5layer CNN model (Table.~\ref{tab:cf10}) have stopping criteria set to 100 epochs.
We decay the step size by $10\times$ in multiple steps at $50^{th}$ and $75^{th}$ epoch. Table~\ref{tab:5cnn-hp} presents the $\alpha$, $\beta$, $\eta$, and $\gamma$ corresponding to the \textit{NGC} mixing weight, momentum, step size, and gossip averaging rate. For all the experiments, we use a mini-batch size of 32 per agent. The stopping criteria is a fixed number of epochs. We have used Nesterov momentum of 0.9 for all CGA and \textit{NGC} experiments whereas D-PSGD and \textit{NGC} with $\alpha=0$ have no momentum.  

\textbf{Hyper-parameters for CIFAR-10 on VGG-11 and ResNet-20:} 
All the experiments for the CIFAR-10 dataset trained on VGG-11 and ResNet-20 architectures (Table.~\ref{tab:cf10}) have stopping criteria set to 200 epochs.
We decay the step size by $10\times$ in multiple steps at $100^{th}$ and $150^{th}$ epoch. Table~\ref{tab:resnet-hp} presents the $\alpha$, $\beta$, $\eta$, and $\gamma$ corresponding to the ngc mixing weight, momentum, step size, and gossip averaging rate. For all the experiments, we use a mini-batch size of 32 per agent. 

{ \textbf{Hyper-parameters used for Table.~\ref{tab:datasets}:}
All the experiments in Table.~\ref{tab:datasets} have stopping criteria set to 100 epochs.
We decay the step size by $10\times$ in multiple steps at $50^{th}$ and $75^{th}$ epoch. Table~\ref{tab:table2-hp} presents the $\alpha$, $\beta$, $\eta$, and $\gamma$ corresponding to the ngc mixing weight, momentum, step size, and gossip averaging rate. For all the experiments related to Fashion MNIST and Imagenette (low resolution of $(32\times 32)$), we use a mini-batch size of 32 per agent. For all the experiments related to CIFAR-100, we use a mini-batch size of 20 per agent.}
\begin{table}[t]
\caption{{Hyper-parameters used for Table.~\ref{tab:datasets}}}
\label{tab:table2-hp}
\begin{center}
\begin{tabular}{c c c c c}
\hline
&Agents & Fashion MNIST & CIFAR-100 & Imagenette\\
\cline{3-5}
Method & (n) & ($\alpha$, $\beta$, $\eta$, $\gamma$) & ($\alpha$, $\beta$, $\eta$, $\gamma$) & ($\alpha$, $\beta$, $\eta$, $\gamma$)\\
 \hline
     D-PSGD &  5 & ($-,0.0,0.01,1.0$)&($-,0.0, 0.1, 1.0$)& ($-,0.0, 0.1, 1.0$)\\
     &  10 & ($-,0.0,0.01,1.0$)&($-,0.0, 0.1, 1.0$)& ($-,0.0, 0.1, 1.0$)\\
  \hline
 \textit{NGC} (ours) &  5 & ($0.0,0.0,0.01,1.0$)&($0.0,0.0, 0.1, 1.0$) & ($-,0.0, 0.1, 1.0$)\\
    ($\alpha=0$) &  10 & ($0.0,0.0,0.01,1.0$)& ($0.0,0.0, 0.1, 1.0$) & ($-,0.0, 0.1, 1.0$)\\
\hline
    CGA &  5 & ($-,0.9,0.01,1.0$)&($-,0.9, 0.1, 1.0$) &($0.0,0.0,0.01,0.5$)\\
      &  10 & ($-,0.9,0.01,1.0$)&($-,0.9, 0.1, 0.5$)  & ($0.0,0.0,0.01,0.5$)\\
     \hline
\textit{NGC} (ours) &  5 & ($1.0,0.9,0.01,1.0$)& ($1.0,0.9, 0.1, 1.0$) &($0.0,0.0,0.01,0.5$)\\
     &  10 & ($1.0,0.9,0.01,1.0$)& ($1.0,0.9, 0.1, 0.5$)  &($0.0,0.0,0.01,0.5$)\\
\hline
   CompCGA &   5 & ($-,0.9,0.01,0.1$)&($-,0.9, 0.01, 0.1$) &($0.0,0.0,0.01,0.1$)\\
    &  10 & ($-,0.9,0.01,0.1$)& ($-,0.9, 0.01, 0.1$)&($0.0,0.0,0.01,0.5$)\\
 \hline
 \textit{CompNGC} (ours) &  5 & ($1.0,0.9,0.01,0.1$)& ($1.0,0.9, 0.01, 0.1$) &($0.0,0.0,0.01,0.1$)\\
    & 10 & ($1.0,0.9,0.01,0.1$)& ($1.0,0.9, 0.01, 0.1$)&($0.0,0.0,0.01,0.5$)\\
 \hline
 \end{tabular}
\end{center}
\end{table}

{\textbf{Hyper-parameters used for Table.~\ref{tab:nlp-fr_imagenette} (right):}
All the experiments in Table.~\ref{tab:nlp-fr_imagenette} have stopping criteria set to 100 epochs.
We decay the step size by $10\times$ at $50^{th}, 75^{th}$ epoch. Table~\ref{tab:imagenette-hp} presents the $\alpha$, $\beta$, $\eta$, and $\gamma$ corresponding to the \textit{NGC} mixing weight, momentum, step size, and gossip averaging rate. For all the experiments, we use a mini-batch size of 32 per agent.
}

\begin{table}[t]
\caption{{Hyper-parameters used for Table.~\ref{tab:nlp-fr_imagenette}  (right) }}
\label{tab:imagenette-hp}
\begin{center}
\begin{tabular}{l  c c }
\hline
&   Ring topology  & Chain topology \\
Method &  ($\alpha$, $\beta$, $\eta$, $\gamma$) & ($\alpha$, $\beta$, $\eta$, $\gamma$) \\
 \hline
 \hline
    D-PSGD & ($-, 0.0, 0.01, 1.0$) &($-, 0.0, 0.01, 1.0$)  \\
    \textit{NGC} $\alpha=0$ (ours)&  ($0.0, 0.0, 0.01, 1.0$)  &($0.0, 0.0, 0.01, 1.0$)\\
      \hline
    CGA &  ($-, 0.9, 0.01, 0.5$)&($-, 0.9, 0.01, 0.1$) \\
    \textit{NGC} & ($1.0, 0.9, 0.01, 0.5$)&($1.0, 0.9, 0.01, 0.1$)\\
      \hline
    CompCGA &  ($-, 0.9, 0.01, 0.1$)&($-, 0.9, 0.01, 0.1$) \\
    \textit{CompNGC} & ($1.0, 0.9, 0.01, 0.1$)  &($1.0, 0.9, 0.01, 0.1$)\\
      
     \hline
\end{tabular}
\end{center}
\end{table}

{\textbf{Hyper-parameters used for Table.~\ref{tab:nlp-fr_imagenette} (left):}
All the experiments in Table.~\ref{tab:nlp-fr_imagenette} (left) have stopping criteria set to 3 epochs.
We decay the step size by $10\times$ at $2^{nd}$ epoch. Table~\ref{tab:nlp-hp} presents the $\alpha$, $\beta$, $\eta$, and $\gamma$ corresponding to the ngc mixing weight, momentum, step size and gossip averaging rate. For all the experiments, we use a mini-batch size of 32 per agent on AGNews dataset. }

\begin{table}[t]
\caption{{Hyper-parameters used for Table.~\ref{tab:nlp-fr_imagenette} (left) }}
\label{tab:nlp-hp}
\begin{center}
\begin{tabular}{l  c c c }
\hline
&  \multicolumn{2}{c}{BERT\textsubscript{mini} }&  DistilBERT\textsubscript{base}\\
Method&   Agents = 4 &Agents = 8&   Agents = 4  \\
&  ($\alpha$, $\beta$, $\eta$, $\gamma$) & ($\alpha$, $\beta$, $\eta$, $\gamma$) &  ($\alpha$, $\beta$, $\eta$, $\gamma$)\\
 \hline
 \hline
    D-PSGD & ($-, 0.0, 0.01, 1.0$) &($-, 0.0, 0.01, 1.0$) & ($-, 0.0, 0.01, 1.0$) \\
    \textit{NGC} $\alpha=0$ (ours)&  ($0.0, 0.0, 0.01, 1.0$)  &($0.0, 0.0, 0.01, 1.0$)& ($0.0, 0.0, 0.01, 1.0$)\\
      \hline
    CGA &  ($-, 0.9, 0.01, 0.5$)&($-, 0.9, 0.01, 0.5$)&($-, 0.9, 0.01, 0.5$) \\
    \textit{NGC} & ($1.0, 0.9, 0.01, 0.5$)&($1.0, 0.9, 0.01, 0.5$)&($1.0, 0.9, 0.01, 0.5$) \\
      \hline
    CompCGA &  ($-, 0.9, 0.01, 0.5$)&($-, 0.9, 0.01, 0.5$)&($-, 0.9, 0.01, 0.5$) \\
    \textit{CompNGC} & ($1.0, 0.9, 0.01, 0.5$)  &($1.0, 0.9, 0.01, 0.5$) & ($1.0, 0.9, 0.01, 0.5$)\\
      
     \hline
\end{tabular}
\end{center}
\end{table}

\textbf{Hyper-parameters for Figures:}
The simulations for all the figures are run for 300 epochs. We scale the step size by a factor or $0.981$ after each epoch to obtain a smoother curve. All the experiment in Figure~\ref{fig:figures}, \ref{fig:alpha}, \ref{fig:norms}, \ref{fig:figures_10agents} and \ref{fig:norms_10agents} use 5 layer CNN network. Experiments in Figure~\ref{fig:figures}, and \ref{fig:norms} use 5 agents while the experiments in Figure~\ref{fig:alpha}, \ref{fig:figures_10agents} and \ref{fig:norms_10agents} use 10 agents. The hyper-parameters for the simulations for all the plots are mentioned in Table~\ref{tab:fig-hp}

\begin{table}[t]
\caption{Hyper-parameters used for Figures~\ref{fig:figures}, \ref{fig:alpha}, \ref{fig:norms}, \ref{fig:figures_10agents} and \ref{fig:norms_10agents}.}
\label{tab:fig-hp}
\begin{center}
\begin{tabular}{|c|c|c|c|c|}
\hline
Figure & $\alpha$ &$\beta$ &$\eta$ & $\gamma$ \\
\hline
\ref{fig:iid_curve} (skew=0,  \textit{NGC})  & 1.0 &0.9 & 0.01 & 1.0\\
\hline
\ref{fig:noniid_curve} (skew=1, \textit{NGC})  & 1.0 &0.9 & 0.01 & 0.1 \\
\hline
\ref{fig:algo_curve} and \ref{fig:norms} (skew=1, ring topology)  & 1.0 &0.9 & 0.01 & 0.1\\
\hline
& 0.0 & &  & 1.0\\
\ref{fig:alpha} (skew=1, \textit{NGC}) & 0.25 & &  & 1.0\\
ring topology & 0.5 & 0.9& 0.01 & 0.5\\
 & 0.75 & &  & 0.5\\
& 1.0 & &  & 0.25\\
\hline
\ref{fig:iid_curve_10agents} (skew=0, \textit{NGC})  & 1.0 &0.9 & 0.1 & 1.0\\
\hline
\ref{fig:noniid_curve_10agents} (skew=1, \textit{NGC})  & 1.0 &0.9 & 0.1 & 0.5\\
\hline
\ref{fig:algo_curve_10agents} (skew=1, ring topology)  & 1.0 &0.9 & 0.1 & 0.5 \\
\hline
 \end{tabular}
\end{center}
\end{table}


\end{document}